\pdfoutput=1
\documentclass[11pt]{article}

\usepackage[]{acl}

\usepackage{times}
\usepackage{latexsym}

\usepackage[T1]{fontenc}
\usepackage[utf8]{inputenc}

\usepackage{microtype}

\usepackage{subfigure}
\usepackage{graphicx}
\usepackage{tabularx} 
\usepackage{float}

\usepackage{inconsolata}
\usepackage{comment}
\usepackage{multirow}
\usepackage{booktabs} 
\usepackage{ulem}
\usepackage{listings}
\usepackage{tabularx}
\usepackage{enumitem}
\title{LLMs Learn Task Heuristics from Demonstrations: A Heuristic-Driven Prompting Strategy for Document-Level Event Argument Extraction}
\renewcommand{\thefootnote}{\fnsymbol{footnote}}
\author{Hanzhang Zhou$^{1,2}$, Junlang Qian$^{1}$, Zijian Feng$^{1,2}$, Hui Lu$^{1}$, Zixiao Zhu$^{1,2}$,
        Kezhi Mao$^{1,2}$\footnotemark[1]
        \\$^{1}$Nanyang Technological University, Singapore\\
        $^{2}$Future Resilient Systems, Singapore-ETH Centre, Singapore\\
  \texttt{\{hanzhang001, junlang001, feng0119, hui007, zixiao001\}@e.ntu.edu.sg} \\
  \texttt{ekzmao@ntu.edu.sg}\\}

\begin{document}
\maketitle
\footnotetext[1]{Corresponding author.}
\renewcommand{\thefootnote}{\arabic{footnote}}
\begin{abstract} 
In this study, we explore in-context learning (ICL) in document-level event argument extraction (EAE) to alleviate the dependency on large-scale labeled data for this task. We introduce the Heuristic-Driven Link-of-Analogy (HD-LoA) prompting tailored for the EAE task. Specifically, we hypothesize and validate that LLMs learn task-specific heuristics from demonstrations in ICL. Building upon this hypothesis, we introduce an explicit heuristic-driven demonstration construction approach, which transforms the haphazard example selection process into a systematic method that emphasizes task heuristics. Additionally, inspired by the analogical reasoning of human, we propose the link-of-analogy prompting, which enables LLMs to process new situations by drawing analogies to known situations, enhancing their performance on unseen classes beyond limited ICL examples. Experiments show that our method outperforms existing prompting methods and few-shot supervised learning methods on document-level EAE datasets. Additionally, the HD-LoA prompting shows effectiveness in other tasks like sentiment analysis and natural language inference, demonstrating its broad adaptability \footnote{Our code is available at \href{https://github.com/hzzhou01/HD-LoA-Prompting}{https://github.com/hzzhou01/HD-LoA-Prompting}}.
\end{abstract}

\section{Introduction}
Document-level Event Argument Extraction (EAE) aims to transform unstructured event information from documents into structured formats encapsulating event arguments, facilitating their interpretation and application in various domains \citep{grishman2019twenty}. The prevalent approach for this task relies on the collection of labeled data and the subsequent model training via supervised learning \citep{ren-etal-2023-retrieve, liu-etal-2023-document, pouran-ben-veyseh-etal-2022-document, zhou-mao-2022-document, du-cardie-2020-document}. While effective, this approach comes with the significant drawback: it necessitates a substantial amount of training data, which is particularly burdensome and costly given the complexity inherent to document-level EAE.

\begin{figure}
    \centering
    {
    \includegraphics[scale=0.135]{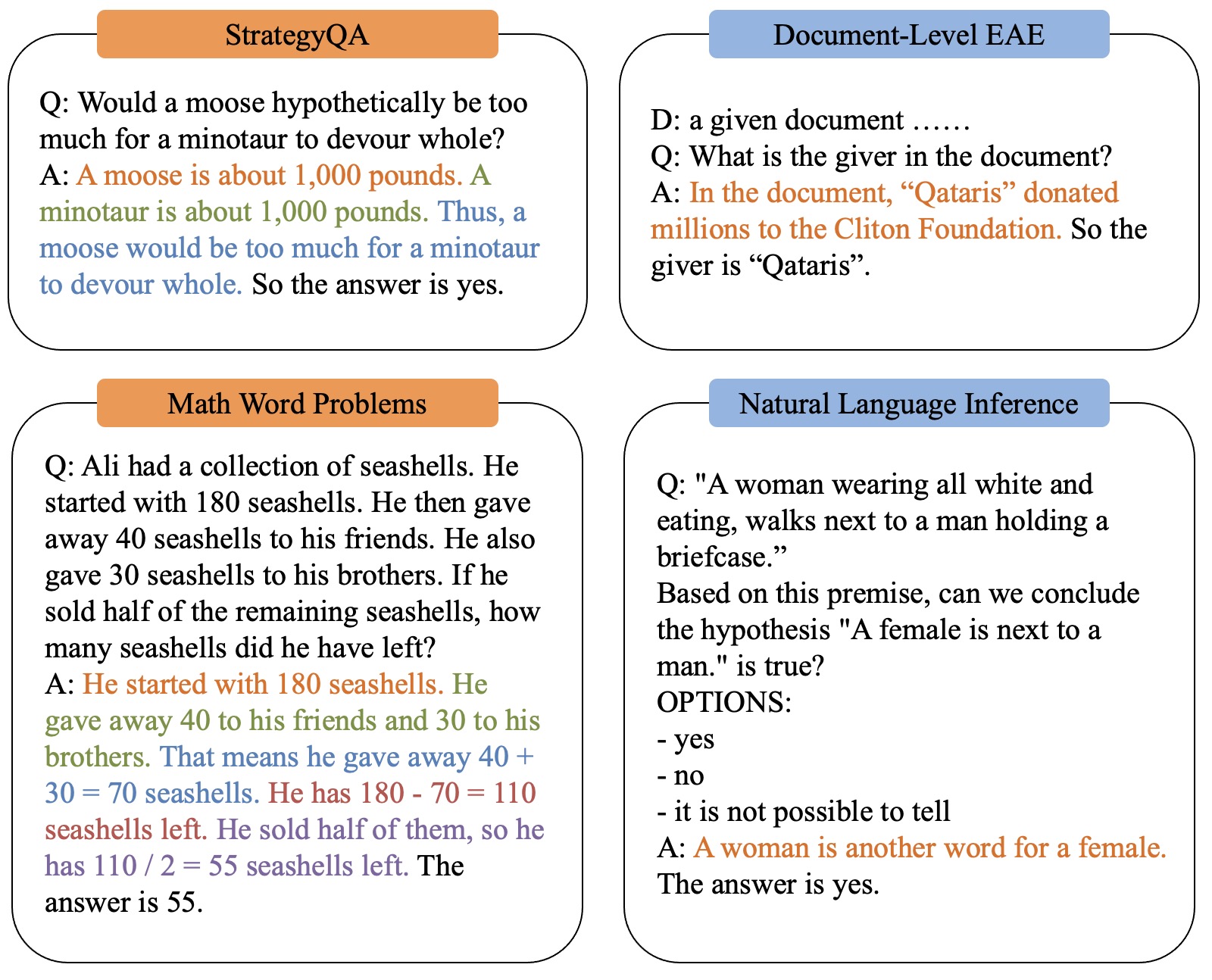}
    }
    \caption{CoT's step-by-step reasoning degrades to a single step for non-reasoning tasks. Reasoning steps of reasoning tasks (in orange) and non-reasoning tasks (in blue) are compared. Different colors indicate distinct reasoning steps. Prompts are from \citep{shum-etal-2023-automatic}.}
    \label{CoT}
\end{figure}

In this context, in-context learning (ICL) \citep{brown2020language, liu-etal-2022-makes,zhou2022least}, an emergent ability of large language models (LLMs), offers a promising alternative to supervised learning. ICL alleviates the need for large-scale data as it only uses a few examples as input-output pairs of the prompt to guide LLMs in performing the task on an unseen example. \\
However, applying ICL to document-level EAE presents numerous challenges. The ICL performance is highly sensitive to the design of in-context demonstrations, such as the selection of examples and the formatting of reasoning steps \citep{zhang2023automatic, zhang-etal-2022-active, fu2022complexity}. Consequently, several crucial challenges emerge concerning the prompting strategy:\\
% Even minor changes in these aspects can lead to substantial performance decline \citep{zhang2023automatic, liu-etal-2022-makes, zhang-etal-2022-active, fu2022complexity}. Consequently, several crucial challenges emerge concerning the prompting strategy:
\textbf{1) Example Selection Challenge.} Selecting optimal in-context examples for ICL is pivotal, yet the understanding of what LLMs learn from these examples remains largely under-explored \citep{wang-etal-2023-label, dong2022survey}. This gap leads to a lack of systematic guidelines, resulting in a disorganized and inefficient example selection process.\\
% This lack of understanding makes it hard to distinguish what are the optimal examples and result in a complicate and chaotic process of example selection. complicates the example selection process, underscoring the need for a deeper investigation into the guiding principles for example selection. 
% and the characteristics that constitute a 'good' example for document-level EAE.
\textbf{2) Context Length Limit}. In document-level EAE, selecting multiple documents as ICL examples could significantly extend the context length, potentially surpassing the token limit of LLMs.\\
\textbf{3) Abundance of Event Types.} The EAE task can involve more than a hundred distinct event types and argument roles. Yet, ICL examples can only capture a narrow subset, leaving the majority of argument roles unseen. Handling unseen classes beyond limited ICL examples is a common problem in classification tasks with diverse class types.\\
% Consequently, it leads to the question: How to design a universal prompting strategy that effectively addresses a wide range of unseen event types and argument roles?
\textbf{4) Prompting Strategy for Non-reasoning Task.} While the chain-of-thought (CoT) prompting is extensively used across a variety of tasks, its effectiveness is compromised in non-reasoning scenarios. As shown in Figure \ref{CoT}, applying CoT to non-reasoning tasks will degrade its step-by-step reasoning into an potentially inadequate single-step. Consequently, there is a need for prompting strategy tailored for non-reasoning tasks.

\begin{figure}
    \centering
    {
    \includegraphics[scale=0.34]{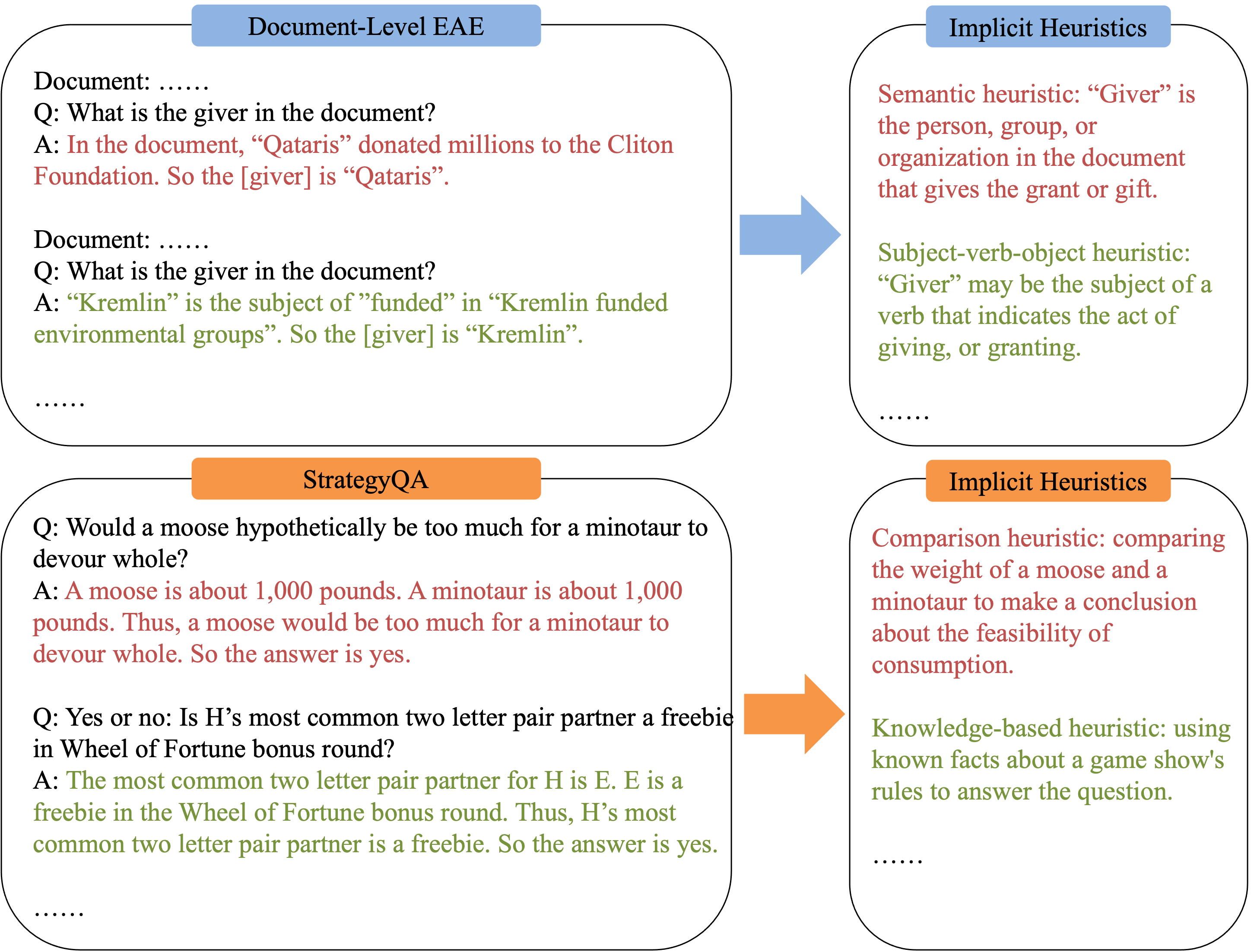}
    }
    \caption{Heuristics are implicitly embedded within explanations of in-context examples.}
    \label{heuristic}
\end{figure}
In this work, we put forward a novel hypothesis that LLMs learn task-specific heuristics from examples and validate it through experiments. Building upon this hypothesis, we propose heuristic-driven link-of-analogy prompting to address the aforementioned challenges. To elaborate:\\
% What LLMs learn from ICL demonstrations remains largely under-explored \citep{wang-etal-2023-label, dong2022survey}, posing challenges to the selection of effective in-context examples.
\textbf{We propose and empirically validate the hypothesis that LLMs learn task specific heuristics from examples in ICL}. Heuristics, defined as \textit{'a high-level rule or strategy for inferring answers to a specific task'}, play a crucial role in human cognition. Humans use heuristics as efficient cognitive pathways, which often lead to more accurate inferences than complex methods \citep{gigerenzer2011heuristic, hogarth2007heuristic}. Similarly, in supervised machine learning (ML) systems, models also learn task-specific patterns through training \citep{shachaf2021theoretical, najafabadi2015deep}. Drawing a parallel, we hypothesize that LLMs learn task-specific heuristics from explanations of in-context examples to aid inference. We qualitatively illustrate how heuristics are \textit{implicitly} embedded in explanations of in-context examples in Figure \ref{heuristic}, and quantitatively validates our hypothesis with experiments detailed in Section \ref{valid_hypo}. \\
Notably, while drawing parallels to supervised ML, ICL is fundamentally different from supervised ML in mechanism: supervised ML learns and updates model parameters during training, whereas LLMs do ICL with all parameters frozen. Therefore, understandings of supervised ML systems (e.g. pattern learning) are not applicable for ICL \citep{min-etal-2022-rethinking, akyurek2022learning}, which necessitates distinct explorations on the mechanism of ICL.
\\
% The hypothesis is derived by drawing an analogy from the training of deep learning models and the fine-tuning of pretrained language models (PLMs). Both these models achieve enhanced performance by learning task-specific patterns, either through training or fine-tuning on task data \citep{zhou-srikumar-2022-closer, shachaf2021theoretical, kim2018interpretability, najafabadi2015deep}. Drawing a parallel, we hypothesize that, although LLMs' parameters are frozen in ICL, providing them with in-context examples allows LLMs to learn task-specific patterns during inference. These patterns encompasses both lower-level patterns like label space and input text distribution, and higher-level heuristics from demonstrations. We notice the learning of lower-level patterns has been validated \cite{min-etal-2022-rethinking}. Therefore, we focus on substantiating the learning of task-specific heuristics in ICL. An illustration of the implicit heuristics behind in-context examples can be found in Figure \ref{heuristic}, while quantitative validations of our hypothesis are detailed in Section \ref{valid_hypo}. \\
\textbf{We propose a heuristic-driven demonstration construction method.} Based on our hypothesis, task heuristics are crucial for the ICL performance of LLMs, yet they are often \textit{implicitly} conveyed through examples. This implicitness complicates the examination of whether demonstrations contain diverse heuristics and leads to uncertainty about whether LLMs have recognized these heuristics. Furthermore, the selection of in-context examples remains an underexplored challenge for ICL. To address these issues, in parallel with human's exploitation of explicit heuristics, our method \textit{explicitly} incorporates task heuristics into demonstrations, transforming the haphazard example selection process into a systematic method that emphasizes task heuristics. \\
%the question: Why not explicitly provide these heuristics in demonstrations? \\
%Our proposed heuristic-driven demonstration construction method presents three key benefits: (1) It simplifies the two-stage process of implicit heuristic recognition and heuristic-based inference in LLMs into a singular heuristic-based inference step, reducing the complexity of the inference process; (2) It addresses the prevailing uncertainty in ICL example selection, converting a directionless and indiscriminate process into a methodical approach that emphasizes task-specific heuristics; (3) It condenses extended sequences of examples that consist of input-output pairs into concise heuristics, mitigating the increase in context length associated with accumulating EAE document examples.\\
\textbf{We propose the link-of-analogy prompting method that is suitable for non-reasoning tasks.} To address the aforementioned challenges of abundance of event types in EAE and the limitations of CoT prompting on non-reasoning tasks, we present the link-of-analogy prompting. Inspired by the analogical reasoning––a core mechanism of human cognition, this approach enables LLMs process new situations (new classes) through drawing an analogy to known situations (known classes). Empirical results demonstrate its effectiveness in enhancing the ICL performance for classes not seen in ICL examples.\\
%This limitation of CoT for non-reasoning tasks aligns with human cognition: while we tend to break down complex problems into smaller ones for reasoning tasks, such as arithmetic problems, we seldom employ this strategy for non-reasoning tasks. The linkf-of-analogy prompting boosts the performance of LLMs by facilitating its understanding of new situations through drawing an analogy to familiar situations.
Our contributions are as follows:
\begin{itemize}[leftmargin=10pt]
    \item We introduce a pioneering work to prompting strategies for the document-level EAE, showcasing significant accuracy improvements on two document-level EAE datasets
    %, with enhancements of $4.65\%$ over baseline prompting methods and $9.50\%$ 
    compared to prompting methods and few-shot supervised learning methods.
    \item We investigate what LLMs learn from ICL, and unveil a new insight that LLMs learn task-specific heuristics from ICL examples.
    \item We propose an heuristic-driven demonstration construction approach, tackling the example selection issue with a fresh perspective on task heuristics, facilitating explicit heuristic learning in ICL. Furthermore, we propose the link-of-analogy prompting, which allows LLMs to process new situations by drawing analogies to known situations.
    % , enhancing ICL performance on unseen classes beyond limited ICL examples.
    \item To further evaluate the adaptability of our method, we validate it on the sentiment analysis and natural language inference tasks, achieving notable accuracy enhancements.
\end{itemize}

\section{What do LLMs learn from the demonstration?}
 \label{valid_hypo}
\begin{figure*}
    \centering
    \includegraphics[scale=0.56]{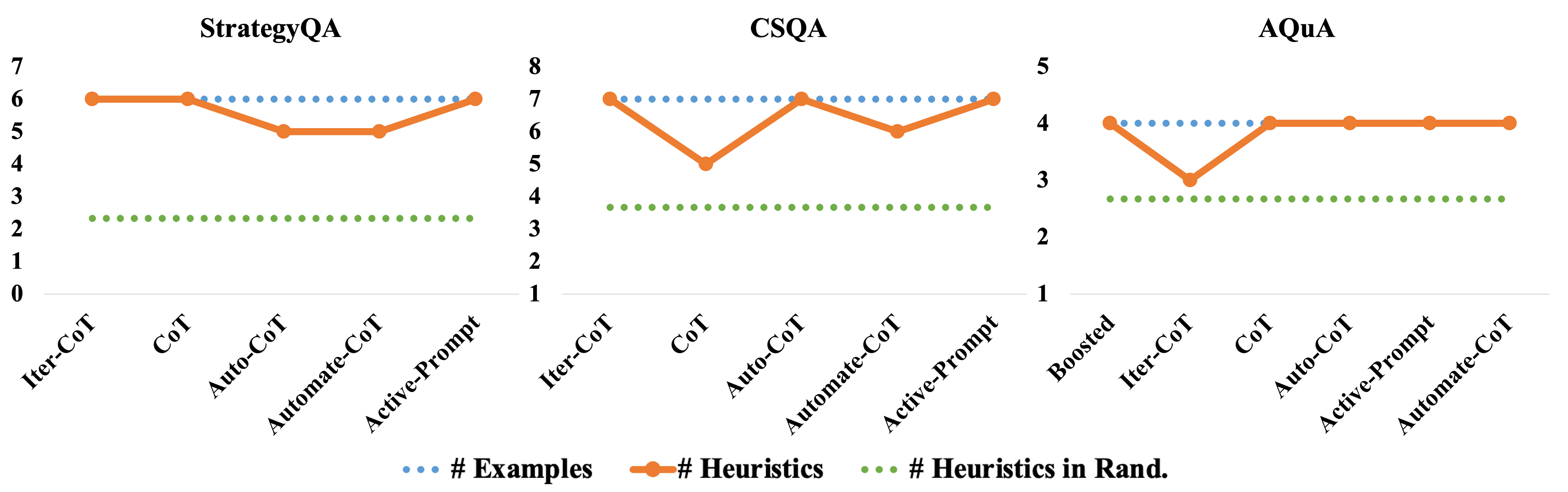}
    \caption{An illustration of the correlation between example quantity and heuristic diversity in well-designed prompts. \# Examples: the number of examples used in each prompt of the corresponding paper. \# Heuristics: the number of heuristics identified in each prompt of the corresponding paper. \# Heuristics in Rand.: the average number of heuristics in the randomly constructed prompt.}
    \label{numbercompare}
\end{figure*}
The understanding of what LLMs learn from the demonstration of ICL remains an open question. In this work, we hypothesize that \textbf{LLMs learn task-specific heuristics from examples in ICL}. We validate this hypothesis with carefully designed experiments in three aspects.
%The hypothesis is derived from inductive reasoning on how deep learning models and pretrained language models learned from the training and fine-tuning, respectively. Specifically, deep learning models improved their performance by learning task-specific patterns during training \citep{kim2018interpretability, najafabadi2015deep, karpathy2015visualizing}. With BERT and similar language models, finetuning has become a more effective approach, as these models extract task-specific patterns from this process\citep{zhou-srikumar-2022-closer, shachaf2021theoretical, howard-ruder-2018-universal}. 
 \subsection{Correlation between Example Quantity and Heuristic Diversity in Well-Designed Prompts}
 Our first experiment operates on the assumption that \textit{if LLMs indeed learn task-specific heuristics from demonstrations, then successful prompts should inherently incorporate a diverse range of heuristics in their examples}, as these heuristics are learnable for LLMs. To examine this proposition, we assess both the quantity of examples and the quantity of different embedded heuristics within prompts from published papers. 
 % If our hypothesis holds, it implies that \textbf{well-constructed prompts should incorporate a diverse range of heuristics}. To validate this implication, we assess both the quantity of examples and the quantity of different heuristics present in prompts from published studies. 
 
% However, manually estimating the number of heuristics can introduce subjectivity, leading to potential biases. Thus, the challenge is 
To objectively identify the number of implicit heuristics embedded in prompts, we employ \texttt{GPT-4} to recognize the embedded heuristic for each example and to determine if it is a shared heuristic across multiple examples. An detailed example of the prompt we used and the heuristics identified by \texttt{GPT-4} can be found in Appendix \ref{heuristics_prompt}.
%In this work, we employ GPT-4 to discern the heuristics underlying each example, and to ascertain if any examples share common heuristics. Precisely, the query for the prompt is, "What is the most critical and profound heuristic at play in each of the following examples? If any two examples share the same heuristic, please indicate this connection. Use no more than two sentences to illustrate the heuristics of each example." Detailed prompts are available in the Appendix.

We investigate the correlation between the number of examples in a prompt and the number of embedded heuristics within the same prompt, analyzing six SOTA prompting methods applied across three distinct datasets. Specifically, prompting methods including CoT \citep{wei2022chain}, Automate-CoT \citep{shum-etal-2023-automatic}, Auto-CoT \citep{zhang2023automatic}, Iter-CoT \citep{sun2023enhancing}, Boosted \citep{pitis2023boosted}, Active-CoT \citep{diao2023active} are investigated and datasets of commonsense reasoning and arithmetic reasoning are evaluated. Our findings in Figure \ref{numbercompare} reveal that: in well-designed prompts, the number of heuristics closely matches the number of examples. Furthermore, the number of heuristics in carefully constructed prompts significantly exceeds that in randomly constructed prompts. This observation substantiates our statement that successful prompts indeed embed a wide array of heuristics in examples.

\begin{figure}
    \centering
    \includegraphics[scale=0.375]{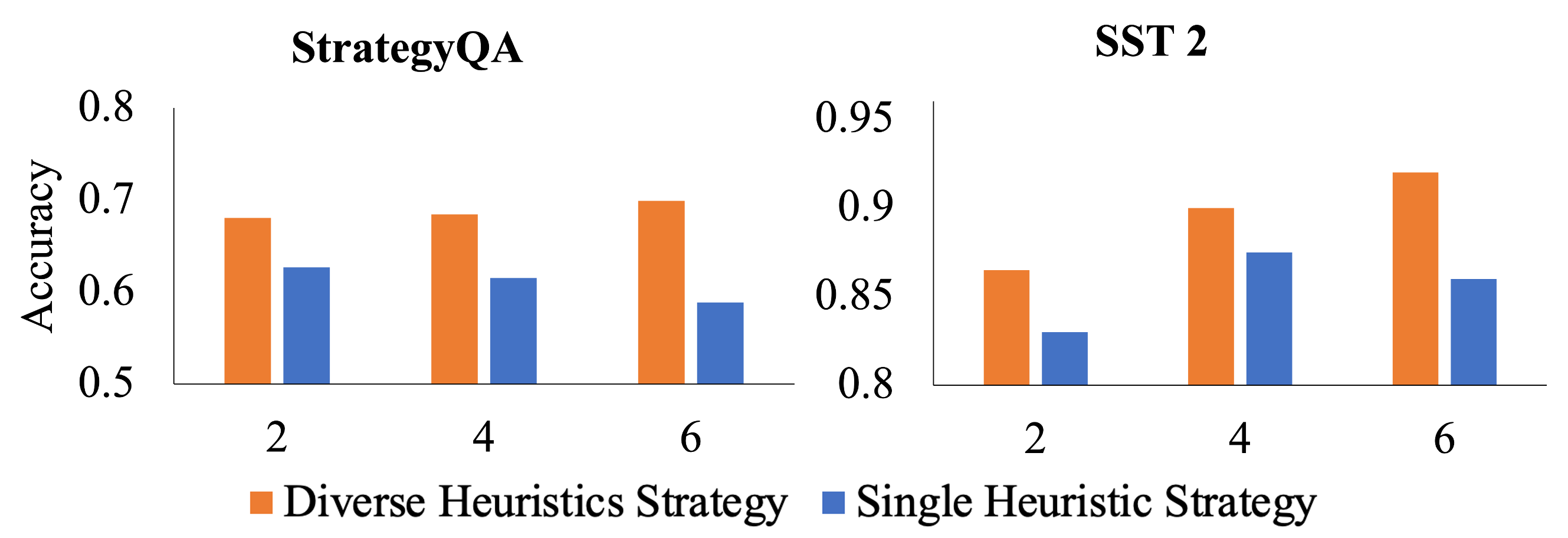}
    \caption{Comparison of ICL performance using single-heuristic strategy versus diverse-heuristics strategy across different number of example on the StrategyQA and SST-2 Dataset.}
    \label{2strategy}
\end{figure}
\begin{figure*}
    \centering
    \label{num1}
    \includegraphics[scale=0.62]{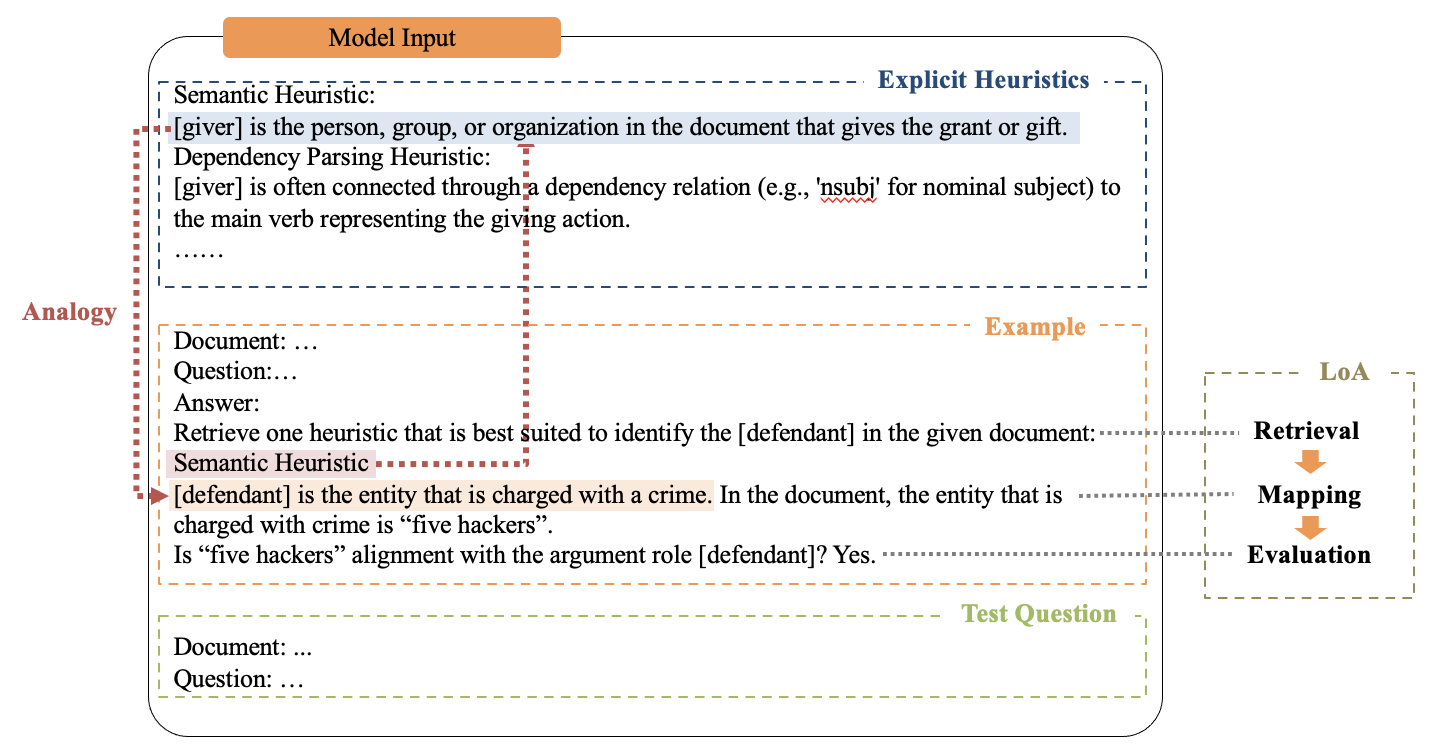}
    \caption{An illustration of HD-LoA prompting.}
    \label{framework}
\end{figure*}
\begin{table*}[h]
\centering
\resizebox{0.9\textwidth}{!}{%
\begin{tabular}{lcccccc}
\hline
 & ER & Comp & KB & Def & Chron & Others \\
\hline
Original Demonstration & 78.5 & 72.7 & 87.2 & 85.1 & 74.7 & 65.5 \\
Heuristic Deduction & 71.4 (\textbf{-7.1}) & 65.4 (\textbf{-7.3}) & 81.6 (\textbf{-5.6}) & 82.9 (\textbf{-2.2}) & 70.3 (\textbf{-4.4}) & - \\
\hline
\end{tabular}
}
\caption{Performance comparison between original demonstration and a demonstration with heuristic deduction (replacing the example of a distinct heuristic type with another example containing a repeated heuristic type).}
\label{table:accuracy_comparison}
\end{table*}
\subsection{Comparing Diverse-Heuristics and Single-Heuristic Strategies}
\label{sec_2strategy}
The second experiment empirically evaluates how the diversity of heuristics within examples impacts ICL performance of the LLM. This experiment is premised under the assumption that \textit{if LLMs \textbf{cannot} learn heuristics from demonstrations, then demonstrations featuring multiple heuristics should yield similar performance with those incorporating a single heuristic}, as heuristics cannot be utilized. To explore this, we compare two distinct example selection strategies. The single-heuristic strategy formulates prompts where all explanations of examples follow a same heuristic. Conversely, the diverse-heuristic strategy constructs prompts where all explanations of examples exhibit different heuristics. We construct prompts that follow these two strategies based on prompts in \citet{diao2023active, shum-etal-2023-automatic}.\\
% If our hypothesis holds true, then incorporating a broader range of heuristics in demonstrations should enable LLMs to identify and leverage more heuristics, leading to enhanced performance. With this in mind, we examine the impact of heuristic diversity in in-context examples on the performance of ICL. Specifically, we juxtapose two in-context example selection strategies on the Strategy QA dataset. The single-heuristic strategy formulates prompts where all rationales of in-context examples follow a same heuristic. Conversely, the diverse-heuristic strategy constructs prompts where all rationales of in-context examples exhibit different heuristics. We construct prompts that follow the two strategies based on the original prompts from \citet{diao2023active, shum-etal-2023-automatic}.\\
\begin{table}[h]
\centering
\resizebox{0.48\textwidth}{!}{%
\begin{tabular}{lcccccc}
\hline
 & ER & Comp & KB & Def & Chron & Others \\
\hline
Count & 14 & 55 & 125 & 47 & 91 & 168 \\
\hline
\end{tabular}
}
\caption{Distribution of samples by heuristic type. "\textit{Others}" includes samples with heuristics not categorized in the predefined types.}
\label{table:heuristic_distribution}
\end{table}
The performance comparison of prompts constructed by the two different strategy on the StrategyQA \citep{geva2021did} and SST-2 \citep{socher-etal-2013-recursive} datasets is illustrated in Figure \ref{2strategy}. The results indicates that, given an equal number of examples, the diverse-heuristics strategy significantly outperforms the single-heuristic approach, which contradicts the assumption. This finding not only validates our hypothesis that LLMs can learn heuristics from in-context examples but also underscores the value of incorporating a variety of heuristics in enhancing ICL performance.
% If LLMs cannot learn heuristics from examples, then both two experiments will have converse results than we reported.
%Given the prompt designed in the paper \citep{diao2023active}, we use text-davinci-003 to generate imitation examples that share the same heuristic for each in-context example. The single-heuristic strategy employs imitation examples derived from a singular example, whereas the diverse-heuristics strategy utilizes imitation examples originating from in-context examples with different heuristics. The performance of the LLM on the StrategyQA \citep{geva2021did} dataset under two different strategies is depicted in Figure \ref{2strategy}. It can be inferred from the results that the diverse-heuristics strategy consistently surpasses the single-heuristic strategy, suggesting that LLMs indeed learn the heuristics from in-context examples. This finding also supports our hypothesis that LLMs indeed learn the heuristics from in-context examples.
\subsection{Impact of Heuristic Deduction Towards ICL Performance}
To validate our hypothesis, we further investigate the impact of reducing an implicit heuristic embedded in demonstration examples. If the classification accuracy of test samples corresponding to this heuristic decreases accordingly, we can validate that LLMs learn heuristics from demonstrations.

We use 500 test samples from the StrategyQA \citep{geva2021did} dataset and the prompt from \citet{shum-etal-2023-automatic} for evaluation. As discussed in Section \ref{sec_2strategy}, we use \texttt{GPT-4} to identify all implicit heuristics embedded in examples of the demonstration: empathetic reasoning (ER), comparison (Comp), knowledge-based (KB), definition-based (Def), and chronological (Chron) heuristics. The prompt for implicit heuristic identification and LLM output are detailed in Appendix \ref{heuristics_prompt}. We then employ \texttt{GPT-4} to label each test sample with the corresponding heuristic that could be used to guide the prediction of the sample. Next, we group the test samples by heuristic type, and the statistics are illustrated in Table \ref{table:heuristic_distribution}. Finally, given the prompt embedded with five different heuristic types, we eliminate the demonstration of a specific heuristic type by replacing its example with another example containing a repeated heuristic type, and monitor the performance change in the corresponding test group. \texttt{gpt-4-1106-preview} is used for evaluation.\\
Experimental results are demonstrated in Table \ref{table:accuracy_comparison}. These results indicate that eliminating the demonstration of a certain heuristic type indeed results in a significant performance drop in the test samples associated with that heuristic, further substantiating our hypothesis that LLMs learn task-specific heuristics from examples. Interestingly, we also observe that samples with heuristics not represented in the demonstration examples (\textit{Others} samples) show significantly lower accuracy, which not only support our hypothesis, but also shed light on example selection, suggesting that selecting examples with their implicit heuristics that cover a wider range of test samples is likely to enhance the ICL performance.

\section{Heuristic-Driven Demonstration Construction}
Building on our understanding of heuristic learning during ICL, we aim to address the challenge of example selection for ICL. Experiments in Section \ref{valid_hypo} indicates that heuristics are crucial for ICL performance of LLMs, yet they are \textit{implicitly} conveyed through explanations of examples. This implicitness complicates the examination of whether ICL demonstrations contain diverse heuristics and leads to uncertainty about whether LLMs have recognized these heuristics. Additionally, when solving a task, humans possess the ability to not only learn from examples but also learn from heuristics for efficient and accurate inference \citep{gigerenzer2011heuristic}. This leads us to question whether LLMs can similarly leverage \textit{explicit} heuristics to improve ICL performance. Therefore, we are motivated to \textit{explicitly} providing LLMs with task-specific heuristics. Our approach is illustrated below:\\
\textbf{Replacing examples with explicit heuristics:} Diverging from traditional prompting strategies that construct prompt with examples where heuristics are implicitly embedded, we propose to replace most examples in the prompt with distinct task-specific heuristics, as demonstrated by the heuristics in Figure \ref{framework}.\\
\textbf{Retaining minimum examples:} A minimal number of examples are preserved to (1) illustrate the formatting of target task and reasoning steps, such as one example is required to illustrate the format of our link-of-analogy prompting, and (2) ensure a balanced coverage of labels in prompt to avoid introducing label bias. Specifically, for document-level EAE task, a single example is maintained to demonstrate the reasoning format. \\
\textbf{Heuristic generation:} A remaining question is how to create the explicit heuristics in the prompt. Both human crafted heuristics and LLM-generated heuristics can be adopted as explicit heuristics. To automate this process, we utilize \texttt{GPT-4} to generate a set of distinct heuristics $S = \{s_1, s_2, ..., s_n\}$ for the document-level EAE task. We adopt $n=10$ in this work. The prompt for heuristic generation and its output are provided in the Appendix \ref{heuristic_generation}. \\
\textbf{Heuristic selection:} Given that not every generated heuristic may suit the target task, we introduce a heuristic selection step. Each heuristic in the generated heuristic set $S$ is individually adopt into a prompt, the ICL performance of each heuristic is evaluated using a subset of the training dataset. Specifically, we employ $1\%$ of the training dataset, identical to the sample size used in the few-shot supervised learning baseline. Through this evaluation, the top-performing heuristics, determined by accuracy, are selected to constitute the explicit heuristic list $\mathbf{H}$ in our prompt. We adopt the top $3$ heuristics in this work. \\
% The number of heuristics selected, denoted as $k$, is similar to the number of examples typically used in conventional prompting strategies. For this study, we set $k=3$. \\
Through this heuristic selection step, low-quality heuristics are excluded. For example, the semantic role labeling heuristic generated in the heuristic generation step (in Appendix \ref{heuristic_generation}) is too specific and of lower quality, thus it demonstrates a significantly low evaluation accuracy ($26.52\%$) compared to a high-quality syntactic heuristic ($33.69\%$).

There are three advantages of our approach. Firstly, it provides a guidance on the example selection process. The example selection process of ICL is often an indiscriminate, manual process \citep{liu2023pre, wei2022chain, zhou2022least}. However, our method converts the directionless and indiscriminate process into a methodical approach that emphasizes task-specific heuristics. Secondly, by emulating human cognitive strategies that leverage explicit heuristics for improved inference—a technique supported by cognitive studies \citep{gigerenzer2011heuristic}—our method enables LLMs to also benefit from heuristic learning during ICL. Finally, it condenses lengthy examples that consists of input-output pairs into compact heuristics, reducing the context length of prompts.

\section{Link-of-Analogy Prompting}
We propose the link-of-analogy prompting to address the challenges below: First, the EAE task is characterized by its extensive variety of argument roles and event types, often exceeding a hundred, yet ICL examples can only cover a very limited subset. This discrepancy raises a critical challenge: designing a prompting strategy that effectively addresses unseen event types. Notably, the issue of handling unseen classes beyond limited ICL examples is a prevalent problem in various NLP tasks. Additionally, to concretize heuristic generation process, we provide heuristics for a specific argument, \textit{giver}, within the prompt. This leads to the question of how to extend \textit{giver} heuristics to other argument roles. Finally, as highlighted in the Introduction, applying CoT prompting to non-reasoning tasks tends to degrade the step-by-step analysis into a one-step rationale \citep{shum-etal-2023-automatic, diao2023active}, necessitating more proper prompting strategies for such tasks.
%CoT prompting \citep{wei2022chain} and its follow-up studies all adopt step-by-step reasoning and they are prevalent prompting method for LLMs. Specifically, chain-of-thought decompose multi-step problems into intermediate steps can  elicit strong reasoning capabilities from LLMs for complex tasks such as math problem. However, CoT prompting is designed for reasoning tasks, which naturally exhibit a step-by-step solving procudure. However, when it comes to non-reasoning tasks, human seldom think step-by-step. It is also reflected by CoT prompts on non-reasoning tasks in \citep{shum-etal-2023-automatic, diao2023active}, the step-by-step format is degraded into a one-step rationale. How to design more proper prompting strategies for non-reasoning tasks?

Inspired by the analogical reasoning \citep{gentner2013analogical}, a core mechanism of human cognition, we seek to resolve the challenges presented. Humans often understand a new situation by drawing an analogy to a familiar situation. For example, students often solve new problems by mapping solutions from known problems \citep{ross1987like}. Similarly, we anticipate that LLMs will be able to extract information of unseen events or generate heuristics for unseen argument roles by drawing an analogy to events and heuristics provided in in-context examples. Empirically, we find that LLMs are indeed capable of doing analogical reasoning when prompted appropriately. For example, when provided with the heuristic for \textit{giver} in the prompt: "\textit{[giver] is the person, group, or organization in the document that gives the grant or gift}", LLMs can make an analogy and generate the heuristic for the argument \textit{vehicle} in the target question: "\textit{[vehicle] is the means of transport used to move the person or object}".

To further enhance the analogical reasoning capabilities of LLMs, we introduce our link-of-analogy (LoA) prompting strategy, which emulates the analogical reasoning process of human. Cognitive science studies reveals that humans perform analogical reasoning through a sequence of \textit{retrieval}, \textit{mapping}, and \textit{evaluation} \citep{gentner2011computational, gentner1997structure}. In alignment with this process, our method involves the same steps.
% retrieving the most appropriate heuristic, mapping this heuristic onto the target argument role to get new heuristic and extracting the argument based on the new heuristic, and finally evaluating the answer. 
Specifically, in the retrieve step, given the base argument role $r_{b}$, a set of heuristics $\mathbf{H} = \{\mathbf{h}_1, \mathbf{h}_2, \cdots, \mathbf{h}_k\}$ for identifying $r_{b}$, a target question and a target argument role $r_{t}$, the LLM will select the most suitable heuristic ${\mathbf{h}_b}$ from $\mathbf{H}$ for identifying $r_{t}$. In the mapping step, the LLM employs analogy mapping $r_{b} : \mathbf{h}_b :: r_{t} : \mathbf{h}_{t}$ to deduce the heuristic $\mathbf{h}_t$ for $r_{t}$. The LLM then infers the argument \( \mathbf{a}_t \) of the target role based on the heuristic $\mathbf{h}_t$.
% \begin{eqnarray}
%     \mathbf{y}_t = \mathop{\arg\max}\limits_{\mathbf{y}_t \in \mathbf{D}} f_{LLM}(\mathbf{y}_t|\mathbf{H}, \mathbf{C}_m, \mathbf{Q}, \mathbf{e}_t, \mathbf{h}_t)
% \end{eqnarray}
% where $\mathbf{D}$ is the document in the target question and $\mathbf{e}_t$ represents the retrieve step.\\
Finally, in the evaluation step, the LLM will reassess the identified argument \( \mathbf{a}_t \). This methodology is exemplified in the in-context example presented in Figure \ref{framework}.

\section{Experiments}
\begin{table*}
\centering\small 
\begin{tabular}{m{2.5cm}<{\centering}m{4.5cm}|m{1cm}<{\centering}m{1cm}<{\centering}|m{2cm}<{\centering}|m{2cm}<{\centering}}
\toprule
\multicolumn{2}{c|}{\multirow{2}{*}{Method}} & \multicolumn{2}{c|}{RAMS}        & DocEE-Normal   & DocEE-Cross    \\
\multicolumn{2}{c|}{}                        & \textbf{Arg-I} & \textbf{Arg-C} & \textbf{Arg-C} & \textbf{Arg-C} \\
\midrule

\multirow{6}{*}{\shortstack{Supervised learning\\(few-shot)}} & EEQA \citep{du-cardie-2020-event}         &  & 19.54 &  &  \\
                             & PAIE \citep{ma-etal-2022-prompt} &     & 29.86 &  &  \\
%                             & DocMRC \citep{liu-etal-2021-machine}      & -     & 45.70 & -     & -     \\
                             & TSAR \citep{xu-etal-2022-two}    & - & 26.67 & -     & -     \\
                             & CRP \citep{liu-etal-2023-document}   &    & 30.09     &  &  \\
                             & FewDocAE \citep{yang-etal-2023-shot}         &  & - & 12.07   & 10.51 \\
\midrule
\multirow{3}{*}{\texttt{text-davinci-003}} & Standard \citep{agrawal-etal-2022-large} & 39.96          & 31.6           & 25.55                & 25.41                \\
                                           & CoT \citep{wei2022chain}                 & 43.03          & 34.94          & 27.68                & 28.64                \\
                                           & HD-LoA (ours)                                   & \textbf{46.17} & \textbf{39.59} & {\textbf{ 30.22}} & {\textbf{31.03}} \\
\midrule

\multirow{3}{*}{\texttt{\shortstack{gpt-3.5\\-turbo-instruct}}}    & Standard \citep{agrawal-etal-2022-large} & 42.44 & 32.46 & 25.67                & 24.48                \\
                                           & CoT \citep{wei2022chain}                 & 40.63          & 33.64          & 26.77                & 25.99                \\
                                           & HD-LoA (ours)                                  & \textbf{43.34}          & \textbf{37.05}           & \textbf{27.98}       & \textbf{27.34}     \\
\midrule
\multirow{3}{*}{\texttt{gpt-4}}         & Standard \citep{agrawal-etal-2022-large} & 44.73               & 37.08                &   29.53              &        27.36         \\
                                        & CoT \citep{wei2022chain}                 & 44.93               & 38.09                & 30.32                & 30.95               \\
                                        & HD-LoA (ours)                                  & \textbf{52.41} & {\textbf{44.12}} &     \textbf{31.53}            &        \textbf{33.48}        \\

\bottomrule
\end{tabular}
\caption{Overall performance. In few-shot setting, the scores of supervised learning methods on RAMS dataset are based on results reported in \citet{liu-etal-2023-document}, where $1\%$ of the training data is used. }
 \label{overall}
\end{table*}

In this section, we aim to explore the following research questions (RQs) regarding our \textbf{H}euristic-\textbf{D}riven \textbf{L}ink-\textbf{o}f-\textbf{A}nalogy (HD-LoA) prompting. \textbf{RQ1} Does HD-LoA prompting improve in-context learning performance in document-level EAE task? \textbf{RQ2} Can HD-LoA prompting effectively mitigate the dependency on extensive labeled data while enhancing accuracy for EAE task? \textbf{RQ3}  Is the HD-LoA prompting effective when applied to tasks beyond EAE? \textbf{RQ4} Do each components of the HD-LoA prompting effectively contributing to its performance?
    % \item \textbf{RQ4} What are the advantages and disadvantages of prompting LLMs compared with finetuning methods in the task of document-level EAE?
\subsection{Experimental Setup}
\textbf{Dataset}: For the evaluation of the document-level EAE task, we adopt RAMS \citep{ebner-etal-2020-multi} and DocEE \citep{tong-etal-2022-docee} datasets.
%RAMS is a commonly used dataset and DocEE distinguishes itself as the most extensive and most recent dataset for document-level EAE tasks. 
The WIKIEVENTS dataset \citep{li-etal-2021-document} is excluded from our study because it relies on preprocessed entity candidates for annotating event arguments the annotation, which diverges from the direct argument identification of LLMs. For evaluation, we follow the metrics in \citep{ma-etal-2022-prompt}, namely the argument identification F1 score (Arg-I), and the argument classification F1 score (Arg-C).\\
Additionally, we utilize the SST-2 \citep{socher-etal-2013-recursive} and SNLI \citep{bowman-etal-2015-large} datasets to assess the effectiveness of our HD-LoA prompting strategy on other non-reasoning tasks: sentiment analysis and natural language inference. The detailed statistics of the datasets and the number of tested samples are listed in Appendix \ref{experimental_detail}.\\
\textbf{Baselines} Our HD-LoA approach is compared against several state-of-the-art prompting methods, including the standard prompting \citep{agrawal-etal-2022-large} used in clinical EAE, and the Chain-of-Thought (CoT) prompting \citep{wei2022chain}. \citet{agrawal-etal-2022-large} presents the only existing method that prompts LLMs in the context of EAE task. Given to its direct question-and-answer format, we refer to it as 'Standard Prompting' in accordance with terminology prevalent in ICL research \citep{wei2022chain}. Notably, as there is no existing prompting strategies tailored for EAE, neither the standard prompting nor CoT prompting has been applied to document-level EAE datasets in the literature. Thus, we report the reproduced results here. \\
Additionally, we compare our method with various supervised learning methods in EAE, such as FewDocAE \citep{yang-etal-2023-shot}, CRP \citep{liu-etal-2023-document}, PAIE \citep{ma-etal-2022-prompt}, TSAR \citep{xu-etal-2022-two}, EEQA \citep{du-cardie-2020-event}, etc. The few-shot comparison results are based on the few-shot performance reported in \citet{liu-etal-2023-document}.  \\
\textbf{LLMs}: The experiments are carried out using three large language models: the publicly available GPT-3 \citep{brown2020language} in its \texttt{text-davinci-003} and \texttt{gpt-3.5-turbo-instruct} versions \citep{ouyang2022training}, as well as \texttt{GPT-4} \citep{OpenAI2023GPT4TR}. Notably, due to the high cost associated with GPT-4, its evaluation is limited to part of the dataset. More experimental details are in Appendix \ref{experimental_detail} and the prompts we used are in Appendix \ref{prompts}.
% These models are accessed via the public APIs from OpenAI’s services\footnote{https://openai.com/api/}.\\
% \textbf{Evaluation Metrics}: We adopt two evaluation metrics following \citep{ma-etal-2022-prompt}: Argument Identification F1 score (Arg-I), and Argument Classification F1 score (Arg-C). 
% \textbf{Experimental Details}: Please find more information about experimental details in Appendix \ref{experimental_detail} and the prompts we used in Appendix \ref{prompts}.

%We adopt two evaluation metrics following \citep{ma-etal-2022-prompt}: Argument Identification F1 score (Arg-I), and Argument Classification F1 score (Arg-C). We provide implementation details and the prompt of our proposed method in the Appendix.
\subsection{Overall Experimental Results}
Addressing \textbf{RQ1}, the experimental results presented in Table \ref{overall} indicate that our HD-LoA prompting significantly enhances in-context learning for document-level EAE task. The HD-LoA method consistently surpasses CoT prompting \citep{wei2022chain} across all three LLMs and both datasets, achieving the largest F1 score improvements of $4.65\%$, $3.41\%$, and $6.03\%$ in Arg-C on each LLM, respectively. In addition, the improvement over the standard prompting \citep{agrawal-etal-2022-large} reaches $7.99\%$ on the \texttt{text-davinci-003} model. 
% However, both CoT and HD-LoA prompting show suboptimal performance on the \texttt{gpt-3.5-turbo} model within the RAMS dataset, possibly due to the model's tradeoff, which favor enhanced dialogue ability at the expense of reasoning ability, potentially impeding the effectiveness of reasoning-based prompting methods.

In response to \textbf{RQ2}, our HD-LoA method, augmented with external knowledge in heuristics, significantly enhances performance in few-shot settings compared to supervised learning approaches. With only one example adopted in the prompt, our HD-LoA achieves a $9.50\%$ F1 score improvement over the CRP method \citep{liu-etal-2023-document} on the RAMS dataset using the \texttt{text-davinci-003} model. Similarly, on the DocEE dataset, our method achieves a substancial $20.52\%$ improvement against FewDocAE \citep{yang-etal-2023-shot}. Experimental findings indicate that our method can successfully mitigate the document-level EAE task's reliance on extensive labeled data while enhancing accuracy.

\begin{table}[h!]
\centering\small
\begin{tabular}{m{3cm}m{1.5cm}m{1.5cm}<{\centering}}
\toprule
& SST-2 & SNLI \\
\midrule
CoT  & 91.39	& 77.97\\
HD-LoA (ours) & \textbf{94.26} & \textbf{80.60} \\
\bottomrule
\end{tabular}
\caption{Evaluation of the HD-LoA prompting on sentiment analysis and natural language inference tasks.}
\label{other_task}
\end{table}

\subsection{Adaptability of HD-LoA Prompting for Other Tasks}
In addressing \textbf{RQ3}, we have extended our HD-LoA prompting method to sentiment analysis (SA) and natural language inference (NLI) tasks, utilizing the SST-2 \citep{socher-etal-2013-recursive} and SNLI \citep{bowman-etal-2015-large} datasets for evaluation. We adopt the CoT style prompts on these two datasets from \citet{shum-etal-2023-automatic}. Experimental results are presented in Table \ref{other_task}. Compared to CoT prompting, our method gets accuracy enhancements of $2.87\%$ and $2.63\%$ on SST-2 and SNLI datasets, respectively. These findings indicate that our HD-LoA prompting can be effectively adapted to a diverse array of non-reasoning NLP tasks. The prompts for SA and NLI tasks are provided in Appendix \ref{prompts}.

% \begin{table*}
% \centering\small 
% \begin{tabular}{m{2.5cm}<{\centering}m{4.5cm}|m{1cm}<{\centering}m{1cm}<{\centering}|m{2cm}<{\centering}|m{2cm}<{\centering}}
% \toprule
% \multicolumn{2}{c|}{\multirow{2}{*}{Method}} & \multicolumn{2}{c|}{RAMS}        & DocEE-Normal   & DocEE-Cross    \\
% \multicolumn{2}{c|}{}                        & \textbf{Arg-I} & \textbf{Arg-C} & \textbf{Arg-C} & \textbf{Arg-C} \\
% \midrule

% \multirow{6}{*}{Supervised learning} & EEQA \citep{du-cardie-2020-event}         & 48.70 & 46.70 & 33.50 & 24.00 \\
%                              & MG-Reader \citep{du-cardie-2020-document} & -     & -     & 32.90 & 21.40 \\
% %                             & DocMRC \citep{liu-etal-2021-machine}      & -     & 45.70 & -     & -     \\
%                              & BART-Gen \citep{li-etal-2021-document}    & 51.20 & 47.10 & -     & -     \\
%                              & OntologyQA \citep{tong-etal-2022-docee}   & -     & -     & 41.00 & 29.80 \\
%                             & PAIE \citep{ma-etal-2022-prompt}          & 56.80 & 52.20 & -     & -    \\
% \midrule
% {\texttt{text-davinci-003}} &  HD-LoA (ours)                  & 46.08 & 39.47 &  30.37 & \textbf{31.21} \\
% \bottomrule
% \end{tabular}
%  \caption{Comparison with Fully Trained Supervised Models.}
%   \label{fully_trained}
%  \end{table*}

\subsection{Comparison with Fully Trained Supervised Models}
We also compare our HD-LoA method with supervised learning method that trained on the entire dataset. It is anticipated that these models trained on thousands of samples would exhibit higher accuracy compared to our method, which employs only a single sample in the prompt. Nevertheless, HD-LoA prompting demonstrates competitive performance against fully trained supervised methods and even outperform these extensively trained models on the DocEE dataset in the cross-domain setting. Experimental results are in Table \ref{fully_trained} of Appendix \ref{fully_supervise}. 

\subsection{Ablations}
\begin{figure}
    \centering
    \includegraphics[scale=0.33]{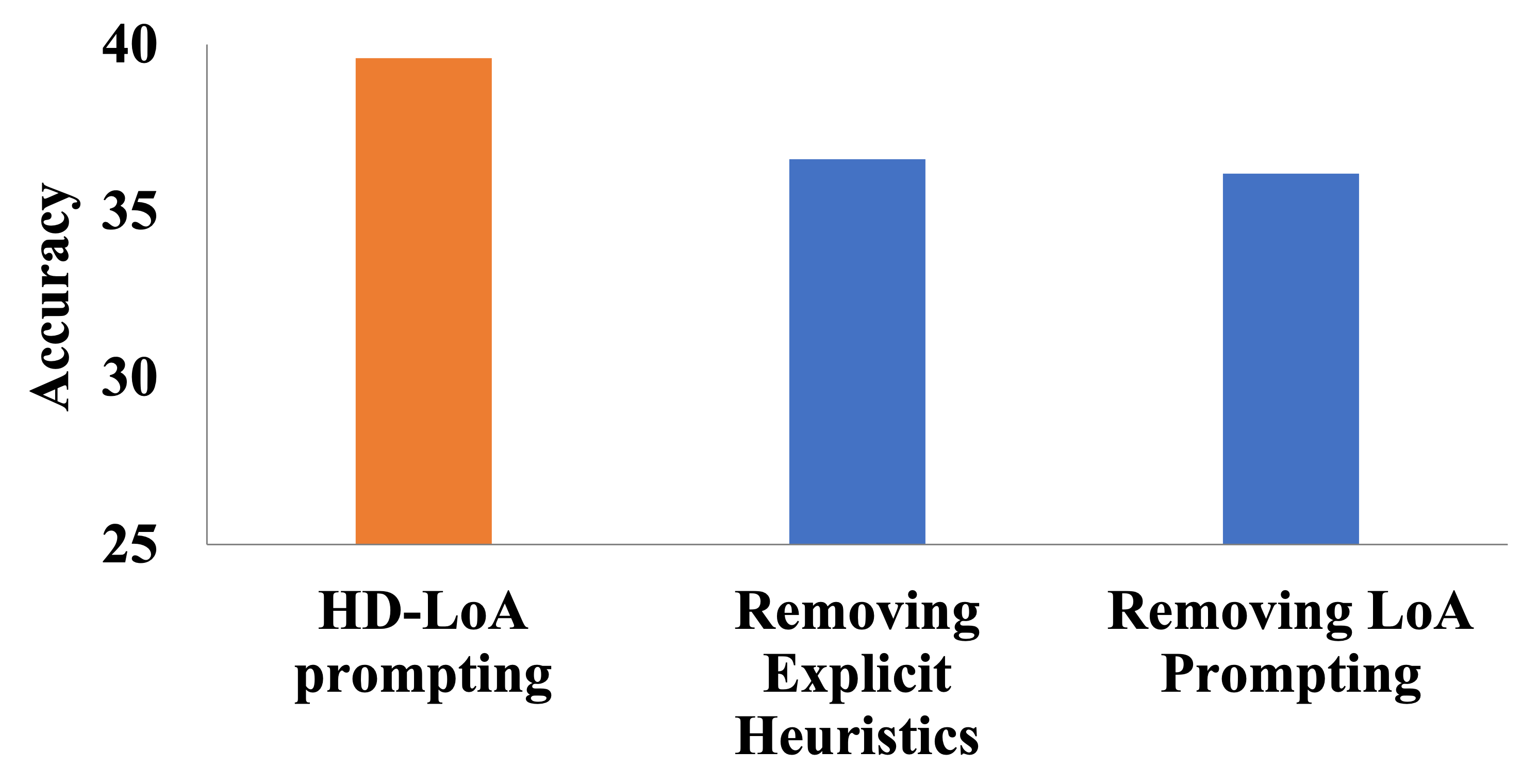}
    \caption{Experimental results of ablations.}
    \label{ablation}
\end{figure}
To address \textbf{RQ4}, we conduct further experiments as follows:
\begin{itemize}[leftmargin=10pt]
    \item \textbf{Ablation Experiments}: We conduct ablation studies on removing the explicit heuristics and removing the link-of-analogy prompting strategy from our prompt. As presented in Figure \ref{ablation}, experimental results on the RAMS dataset demonstrate that removing either the task-specific heuristics or link-of-analogy prompting will significantly degrade the ICL performance of the HD-LoA prompting, suggesting the effectiveness of each component of our prompting strategy.
    % Experimental results detailed in Appendix \ref{Ablations} show effectiveness of each component of our method.
    \item \textbf{Seen Classes and Unseen Classes Accuracy Increase Comparison for LoA}: To further validate the objective of the LoA prompting strategy, which aims to enhance ICL performance for unseen classes in the prompt, we evaluate and compare the accuracy increase facilitated by LoA prompting for both seen and unseen classes in Appendix D. Results detailed in Appendix \ref{Unseen Classes} show that LoA prompting is indeed effective in enhancing ICL performance on classes unseen in the prompt.
    % \item \textbf{Explicit Heuristic Selection}: We compare the ICL performance between heuristics selected by our method and randomly selected heuristics in Appendix.
\end{itemize}
% we carry out ablation experiments to discern how each component contributes to the overall performance of our HD-LoA prompting method. As presented in Table \ref{ablation}, experimental results demonstrate that removing either the task-specific heuristics or link-of-analogy prompting will significantly degrade the eperformance of the HD-LoA prompting, suggesting the importance of each part of our prompting strategy.
%\subsection{Analysis of Fine-tuning and ICL}
%We theoretically analysis the advantanges and disadvantanges of ICL and fine-tuning method in the context of EAE. Firstly, as suggested by \citet{grishman2019twenty}, EAE significantly involves word sense disambiguation. Due to their expansive training corpora, LLMs demonstrate superior capabilities in mitigating such disambiguation. Furthermore, ICL exhibits exceptional proficiency in few-shot learning scenarios. On the contrary, the fine-tuning approach enables learning of more nuanced task-specific patterns, such as the distinctive characteristics of the labeling process. As a result, the fine-tuning method tends to achieve superior accuracy.
\section{Understanding Why HD-LoA Prompting Works}
Following the empirical validation of the effectiveness of our HD-LoA prompting, this section delves into an analysis to elucidate why our method works.\\
\textbf{Analysis of heuristic-driven demonstration construction method}: Firstly, our method naturally incorporates diverse distinct heuristics in prompt. As shown in Section \ref{sec_2strategy}, inclusion of diverse heuristics can significantly boost the ICL performance. In addition, cognitive research finds that humans use heuristics as efficient cognitive pathways to achieve more accurate inferences compared to complex methods \citep{gigerenzer2011heuristic, hogarth2007heuristic}. Paralleling this human cognitive strategy, we enable LLMs to learn from explicit heuristics to enhance inference. Specifically, for LLMs demonstrating suboptimal performance with Standard Prompting and in non-reasoning tasks where definitive rationales are elusive, the provision of explicit heuristics offers LLMs helpful strategies to use and enhance inference. Moreover, as discussed in Section \ref{valid_hypo}, LLMs use implicit heuristics embedded conventional prompts to facilitate inference. By converting these implicit heuristics to explicit heuristics offers a more straightforward way to utilize heuristics and may potentially simplify the utilization of heuristics by LLMs.\\
\textbf{Analysis of the link-of-analogy prompting}: The LoA prompting, which is inspired by the analogical reasoning of human cognition, enables LLMs to process new situations by drawing analogies to known situations. This ability is particularly useful in ICL, where LLMs are always facing unseen samples and unseen classes. As evidenced by experiments in Appendix \ref{Unseen Classes}, the LoA prompting is indeed effective in enhancing ICL performance for classes unseen in the prompt.
\section{Related Work}
\textbf{Document-level EAE} Existing document-level EAE studies are mostly based on supervised learning methods, which relies on the extensive collection of labeled data \citep{hsu-etal-2023-ampere, ma-etal-2022-prompt, pouran-ben-veyseh-etal-2022-document, zhou-mao-2022-document, xu-etal-2021-document, ebner-etal-2020-multi}. Only \citet{agrawal-etal-2022-large} exploits adopting LLMs on clinical EAE though standard prompts that not involve any reasoning strategies. Considering the potential of ICL to reduce the dependency on large-scale labeled datasets and the revolutionize impact of LLMs, it is lack of study on prompting strategy tailored for the EAE task. \\
\textbf{In-context learning} ICL enables LLMs to perform a target task by feeding a few prompted examples as part of the input \citep{brown2020language}. As the mechanism of ICL is fundamentally different from supervised ML, the working mechanism of ICL remains an open question \citep{dong2022survey}. Few studies have conducted preliminary explorations: \citet{min-etal-2022-rethinking} showed that the label space, input text distribution and overall format contribute to the ICL performance. \citet{liu-etal-2022-makes} concluded that examples that are semantically similar to the test sample are more effective. \citet{akyurek2022learning} found that transformer-based ICL can implement standard finetuning implicitly. In this work, we further hypothesize and validate that LLMs learn task-task specific heuristics from examples via ICL.\\
Moreover, the performance of ICL is very sensitive to example selection \citep{gonen2022demystifying} and the optimal selection criteria remains unclear. Various studies proposed different ways: selecting examples based on complexity \citep{fu2022complexity}, mutual information \citep{sorensen-etal-2022-information}, diversity \citep{zhang2023automatic}, labeled dataset \citep{shum-etal-2023-automatic}, etc. In this work, we convert the indiscriminate example selection process into a methodical approach that emphasizes task heuristics, making the example selection process more transparent. \\
% \textbf{Chain-of-thought prompting} CoT prompting \citep{wei2022chain} is a paradigm for prompting LLMs \citep{wang2022self, fu2022complexity, zhang2023automatic}. It enhance the LLM's performance on many reasoning tasks, however, when applying CoT on non-reasoning tasks, due to the different nature of task, the step-by-step reasoning process is degraded to a one-step reasoning process \citep{shum-etal-2023-automatic, wang2022rationale}, thus limits its performance on non-reasoning tasks. 
%Therefore, a more suitable prompting strategy for non-reasoning task is needed.
\section{Conclusion}
In this work, we hypothesize and validate that LLMs learn task-specific heuristics from demonstrations during ICL, which can provide a guidance and simplify the example selection process. Building upon this hypothesis, we introduce an explicit heuristic-driven demonstration construction strategy, and propose a link-of-analogy prompting method. These methods shed light on the heuristic learning of LLMs and the challenge of handling unseen classes in ICL. Extensive experimentation demonstrates the effectiveness and adaptability of our HD-LoA prompting.

\section*{Limitations}
\textbf{Dependency on advanced reasoning abilities of LLMs.} In this work, we aims to explore the upper bounds of in-context learning performance on EAE task in the few-shot setting.
%by enabling LLMs to perform analogical reasoning and explicitly utilize high-level heuristics. 
Our method's reliance on using the sophisticated reasoning capabilities in LLMs makes it unsuitable for models with limited reasoning capabilities. For example, the limited reasoning ability of the \texttt{gpt-3.5-turbo-instruct} model could hinder the performance of our method. However, our findings that LLMs can learn heuristics from in-context examples is applicable to diverse LLMs. \\
\textbf{Heuristic Quality}. The heuristic quality is important for our method. We address this issue by enhancing the probability of generating high-quality heuristics and filtering out low-quality heuristics. We generates an excessive number of heuristic candidates to increase the chances of including high-quality heuristics. Subsequently, we filter out low-quality heuristics by assessing the accuracy of each heuristic candidate on a small set of samples. Future work could explore more sophisticated heuristic generation strategies, such as generating heuristics with diverse granularity or refining heuristics based on feedback from misclassified examples.

\section*{Acknowledgements}
The authors would like to thank Edmond Lo, Lihui Chen, Xiyu Zhang, and the anonymous reviewers for their constructive comments and suggestions. The research was conducted at the Future Resilient Systems at the Singapore-ETH Centre, which was established collaboratively between ETH Zurich and the National Research Foundation Singapore. This research is supported by the National Research Foundation Singapore (NRF) under its Campus for Research Excellence and Technological Enterprise (CREATE) programme.

\bibliography{citations}

\appendix
%\section{Appendix}
\section{Experimental Details}
\label{experimental_detail}
\begin{table*}[h]
\centering\small 
\begin{tabular}{m{4cm}<{\centering}|m{3.5cm}<{\centering}|m{1.5cm}<{\centering}|m{1.5cm}<{\centering}|m{1.5cm}<{\centering}}
\toprule
      Dataset & Task Type &  \# Example & \# Eval. & Eval. Split \\
\midrule
      RAMS \citep{ebner-etal-2020-multi}  & Doc-Level EAE & 1 & 871 & Test\\
      DocEE \citep{tong-etal-2022-docee} & Doc-Level EAE & 1 & 800 & Test \\
      SST-2 \citep{socher-etal-2013-recursive} & Sentiment Analysis & 2 & 872 & Validation \\
      SNLI \citep{bowman-etal-2015-large} & Natural Language Inference & 3 & 500 & Test\\
\bottomrule
\end{tabular}
\caption{The overall statistics of the dataset. \# Example: The number of examples used in the HD-LoA prompting. \# EVAL.: the number of samples used for evaluation of different prompting methods. EVAL. Split: evaluation split.}
 \label{data_statistics}
\end{table*}
The statistics of the dataset are provided in Table \ref{data_statistics}. We use the test split of RAMS dataset and the validation split of SST-2 for evaluation, following the setting in \citet{wang2022rationale}. Considering the extensive size of the DocEE and SNLI datasets, which makes a full-scale evaluation using LLMs impractical, we follow \citet{shum-etal-2023-automatic, wang2022rationale} and evaluate a subset of these datasets. Owing to the substantial costs associated with deploying GPT-4, we restrict its evaluation on the RAMS dataset and DocEE dataset to 200 samples. In addition, regarding the DocEE dataset, it presents two distinct settings. In the conventional configuration, the training and testing data share an identical distribution. Conversely, the cross-domain setup features training and testing data composed of non-overlapping event types. Furthermore, our heuristic-driven demonstration construction method necessitates far fewer examples than traditional prompting methods, only keeping the minimum number of examples to avoid bias in example answers. Specifically, for the EAE task, we use only one example, and for sentiment analysis and natural language inference tasks, two and three examples are employed respectively.

We evaluate our prompting method on \texttt{text-davinci-003}, \texttt{gpt-3.5-turbo-instruct} and \texttt{GPT-4} \citep{OpenAI2023GPT4TR}. The pricing for running these models ranges from USD 0.0015 per 1,000 tokens to USD 0.03 per 1,000 tokens. The \texttt{gpt-3.5-turbo-instruct} model is of the lowest cost but exhibits limited reasoning capabilities. We employ these LLM models from the OpenAI API. During the all experiments, the temperature is fixed as $0$. In the evaluation of document-level EAE datasets, we omit articles from both the ground truth and the prediction during the assessment to align the more closely with the real content of event arguments.

\section{Comparison with Fully Supervised Methods}
\label{fully_supervise}
We compare our HD-LoA method with supervised learning method that trained on the entire dataset for document-level EAE task. As illustrated in Table \ref{fully_trained}, it is anticipated that these models trained on thousands of samples would exhibit higher accuracy compared to our HD-LoA method, which employs only a single labeled sample. Nevertheless, HD-LoA prompting demonstrates competitive performance against supervised methods and even outperform these extensively trained models on the DocEE dataset in the cross-domain setting. This finding also illustrates the effectiveness of our HD-LoA prompting strategy, particularly in scenarios where it is impractical and costly to build large annotated datasets.
\begin{table*}
\centering\small 
\begin{tabular}{m{2.5cm}<{\centering}m{4.5cm}|m{1cm}<{\centering}m{1cm}<{\centering}|m{2cm}<{\centering}|m{2cm}<{\centering}}
\toprule
\multicolumn{2}{c|}{\multirow{2}{*}{Method}} & \multicolumn{2}{c|}{RAMS}        & DocEE-Normal   & DocEE-Cross    \\
\multicolumn{2}{c|}{}                        & \textbf{Arg-I} & \textbf{Arg-C} & \textbf{Arg-C} & \textbf{Arg-C} \\
\midrule

\multirow{6}{*}{Supervised learning} & EEQA \citep{du-cardie-2020-event}         & 48.70 & 46.70 & 33.50 & 24.00 \\
                             & MG-Reader \citep{du-cardie-2020-document} & -     & -     & 32.90 & 21.40 \\
%                             & DocMRC \citep{liu-etal-2021-machine}      & -     & 45.70 & -     & -     \\
                             & BART-Gen \citep{li-etal-2021-document}    & 51.20 & 47.10 & -     & -     \\
                             & OntologyQA \citep{tong-etal-2022-docee}   & -     & -     & 41.00 & 29.80 \\
                            & PAIE \citep{ma-etal-2022-prompt}          & 56.80 & 52.20 & -     & -    \\
\midrule
{\texttt{text-davinci-003}} &  HD-LoA (ours)                  & 46.17 & 39.59 & 30.22 & {\textbf{31.03}} \\
\bottomrule
\end{tabular}
 \caption{Comparison with Fully Trained Supervised Models.}
  \label{fully_trained}
 \end{table*}

% \begin{figure}
%     \centering
%     \includegraphics[scale=0.3]{figures/ablations.png}
%     \caption{Experimental results of ablations.}
%     \label{ablation}
% \end{figure}
\begin{figure}
    \centering
    \includegraphics[scale=0.3]{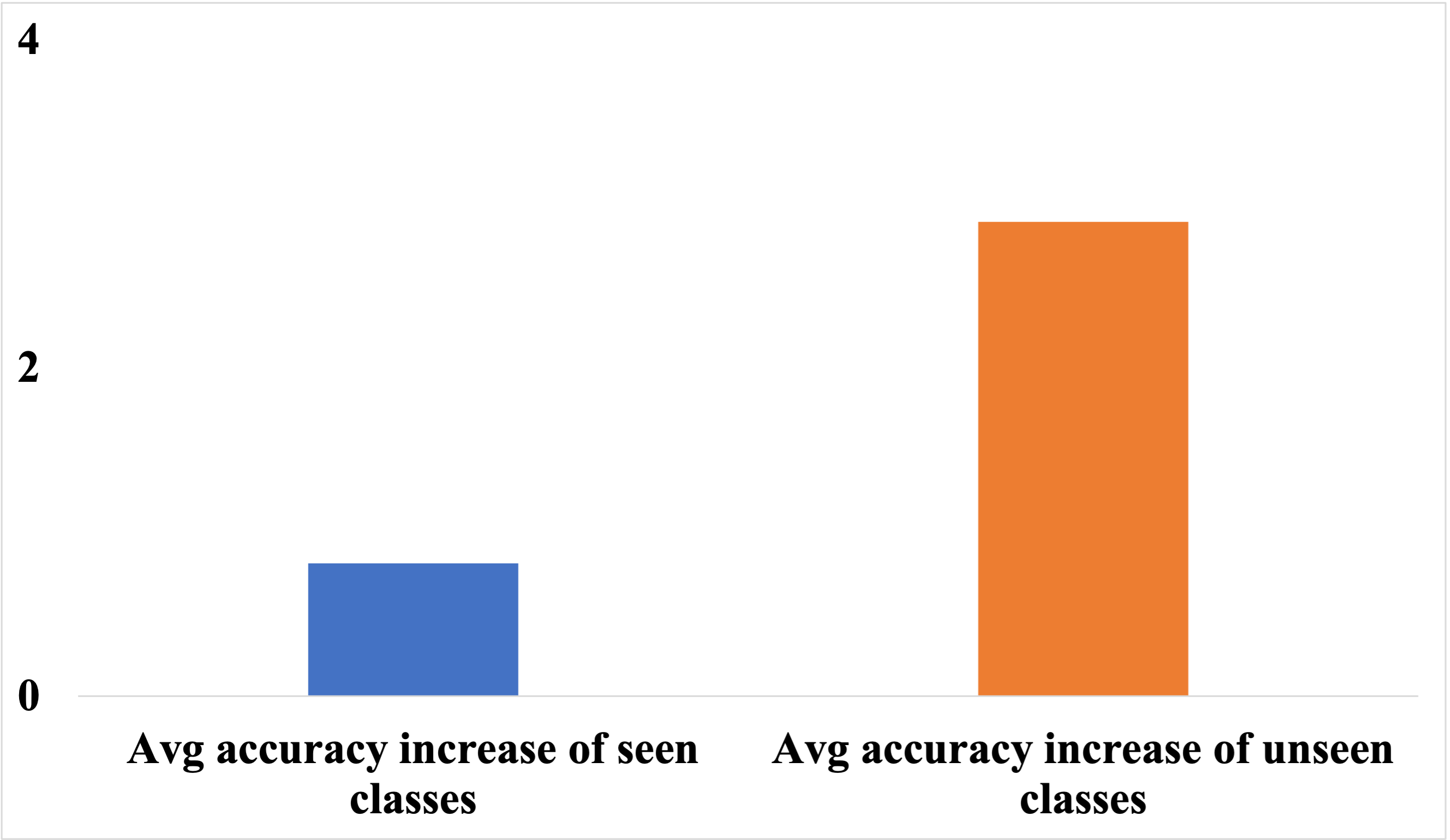}
    \caption{Seen classes and unseen classes accuracy increase comparison with LoA prompting.}
    \label{seen_and_unseen}
\end{figure}
% \section{Ablations}
% \label{Ablations}
% we carry out ablation experiments to discern how each component contributes to the overall performance of our HD-LoA prompting method. As presented in Figure \ref{ablation}, experimental results on the RAMS dataset demonstrate that removing either the task-specific heuristics or link-of-analogy prompting will significantly degrade the ICL performance of the HD-LoA prompting, suggesting the effectiveness of each component of our prompting strategy.

\section{{Seen Classes and Unseen Classes Accuracy Increase Comparison for LoA}} 
\label{Unseen Classes}
We conduct experiment to validate the effectiveness of the LoA prompting strategy in enhancing ICL performance for unseen classes. Given that in-context examples can only capture a narrow subset of classes (seen classes), leaving the majority of argument roles unseen, we assess and compare the accuracy increase of adopting LoA prompting for seen classes against unseen classes. Experimental results in Figure \ref{seen_and_unseen} show that LoA prompting results in a more significant accuracy increase on unseen classes compared to seen classes. It indicates that the LoA prompting is indeed effective in enhancing ICL performance on classes unseen in the prompt.

\section{Recognize Implicit Heuristics of In-Context Examples by GPT-4}
\label{heuristics_prompt}
The prompt we use to recognize implicit heuristics inherent in examples and the output of GPT-4 is given below.
\begin{table}[H]
  \centering
  \begin{tabular}{p{7cm}} % 使用p列类型，指定列宽为6cm
    \toprule
    \multicolumn{1}{>{\fontsize{7}{5}\selectfont\bfseries}l}{Prompt Input} \\
    \midrule
    \begin{lstlisting}
What is the most critical and profound heuristic at play in each of the following examples? If any two examples share the same heuristic, please indicate this connection. Use no more than two sentences to illustrate the heuristics of each example.

example 1
Q: Did the death of Helen Palmer have a significant effect on Dr. Seuss?
A: Helen Palmer was Dr. Seuss's wife. The death of a spouse can have a significant effect on a person. Thus, the death of Helen Palmer could have a significant effect on Dr. Seuss. So the answer is yes.

example 2
Q: Would a moose hypothetically be too much for a minotaur to devour whole?
A: A moose is about 1,000 pounds. A minotaur is about 1,000 pounds. Thus, a moose would be too much for a minotaur to devour whole. So the answer is yes.

example 3
Q: Yes or no: Is H's most common two letter pair partner a freebie in Wheel of Fortune bonus round?
A: The most common two letter pair partner for H is E. E is a freebie in the Wheel of Fortune bonus round. Thus, H's most common two letter pair partner is a freebie in the Wheel of Fortune bonus round. So the answer is yes.

example 4
Q: Yes or no: Would bald eagle deliver an urgent message before B-52?
A: The top speed of a bald eagle is about 80 mph. The top speed of a B-52 is about 650 mph. Thus, a B-52 would be faster than a bald eagle. So the answer is no.

example 5
Q: Yes or no: Does an individual oceanographer study many sciences?
A: Oceanography is the study of the ocean. The ocean is a complex system. Thus, oceanography is a multidisciplinary science. So the answer is yes.

example 6
Q: Yes or no: Could Steven Spielberg send emails as a child?
A: Steven Spielberg was born in 1946. Email was invented in 1971. Thus, Steven Spielberg could not send emails as a child. So the answer is no.
\end{lstlisting} \\
  \bottomrule
  \end{tabular}
\end{table}
\begin{table}[H]
  \centering
  \begin{tabular}{p{7cm}} % 使用p列类型，指定列宽为6cm
    \toprule
    \multicolumn{1}{>{\fontsize{7}{5}\selectfont\bfseries}l}{LLM Output} \\
    \midrule
    \begin{lstlisting}
Example 1: This employs the "empathetic reasoning" heuristic, assuming that because most people would be significantly affected by the death of a spouse, so too would Dr. Seuss.

Example 2: The "comparison heuristic" is used here, comparing the size and weight of a moose to that of a minotaur to make a conclusion about the feasibility of consumption.

Example 3: This uses the "knowledge-based" heuristic, using known facts about a game show's rules to answer the question.

Example 4: This is another instance of the "comparison heuristic," this time comparing the top speeds of a bald eagle and a B-52 to determine which would deliver a message faster.

Example 5: The "definition-based" heuristic is applied here, relying on the understanding of what the field of oceanography entails to conclude that it involves the study of multiple sciences.

Example 6: The "chronological heuristic" is used here, comparing the timelines of Spielberg's childhood and the invention of email to answer the question.
\end{lstlisting} \\
    \bottomrule
  \end{tabular}
\end{table}

\section{Explicit Heuristic Generation by GPT-4}
\label{heuristic_generation}
The prompt we utilize to generate explicit heuristics for document-level EAE and the output by GPT-4 is provided below.
\begin{table}[H]
  \centering
  \begin{tabular}{p{7cm}} % 使用p列类型，指定列宽为6cm
    \toprule
    \multicolumn{1}{>{\fontsize{7}{5}\selectfont\bfseries}l}{Prompt Input} \\
    \midrule
    \begin{lstlisting}
Generate 10 high-level and concise heuristics for identifying the event argument of  'giver' in a document, as part of the event argument extraction task. Each heuristic should offer a distinct perspective and adhere to the format demonstrated in the provided example.

Example heuristic:
Semantic heuristic: The 'giver' is identified as the individual, group, or organization mentioned in the document that is responsible for providing a gift or grant.
\end{lstlisting} \\
  \bottomrule
  \end{tabular}
\end{table}

\begin{table}[H]
  \centering
  \begin{tabular}{p{7cm}} % 使用p列类型，指定列宽为6cm
    \toprule
    \multicolumn{1}{>{\fontsize{7}{5}\selectfont\bfseries}l}{LLM Output} \\
    \midrule
    \begin{lstlisting}
Grammatical heuristic: The 'giver' is typically the subject in a sentence structure where a verb of giving (e.g., 'donate', 'provide', 'offer') is used, indicating the action performer.
Lexical heuristic: The 'giver' can often be identified by keywords or phrases that signify giving, such as 'donor', 'benefactor', or 'sponsor', appearing in close proximity to the action of giving.
Syntactic heuristic: The 'giver' may be recognized by analyzing sentence structure, often appearing before prepositional phrases starting with 'to' that introduce the recipient (e.g., "X gives Y to Z", X is the 'giver').
Semantic role labeling heuristic: In sentences where an action of giving is present, the 'giver' is likely the entity filling the semantic role of 'Agent' or 'Source', initiating the action.
Coreference resolution heuristic: The 'giver' might be determined by identifying pronouns or noun phrases that refer back to an individual, group, or organization previously mentioned as performing the action of giving.
Dependency parsing heuristic: In parsing the sentence structure, the 'giver' is often connected through a dependency relation (e.g., 'nsubj' for nominal subject) to the main verb representing the giving action.
Discourse analysis heuristic: The 'giver' can be inferred from the broader context or narrative of the document, where a party is described or implied to be in a position to give, support, or provide resources.
Named Entity Recognition (NER) heuristic: The 'giver' is often a named entity categorized as a person, organization, or geopolitical entity, involved in a transactional or giving action.
Pragmatic heuristic: The 'giver' may be implied through indirect speech acts or politeness strategies, where the action of giving is not explicitly stated but understood from the context or social norms.
Co-occurrence heuristic: The 'giver' might be associated with frequent co-occurrence with verbs or nouns related to giving in a large text corpus, indicating a statistical likelihood of being the giving party.
\end{lstlisting} \\
    \bottomrule
  \end{tabular}
\end{table}

\section{Full Prompts}
\label{prompts}
\subsection{The Demonstration for SST-2 Dataset}
\begin{lstlisting}
Instruction: You are to perform a 2-class sentiment analysis on the given sentences. For each sentence, determine if the sentiment is either "positive" or "negative". Specifically, you will leverage the pattern in the pattern list below to recognize the sentiment of the target sentence. For each sentence, you will select the most appropriate one or two patterns to identify its sentiment. The identification of a sentence based on different patterns should not be interconnected. Finally, re-evaluate the identified sentiment to get the final answer.

Pattern list:
[
Overall Context Pattern: Examine the general tone of the entire sentence. Determine if it overall express a positive or negative sentiment, regardless of individual words.
Descriptive Phrases pattern: Identify key adjectives or descriptive phrases that convey strong feelings or opinions about the subject.
Negation Handling Pattern: Look for negative words like "not," "didn't," or "never" that might reverse the sentiment of the words following them. Ensure the reversed sentiment is understood correctly.
Comparatives Pattern: Search for words or phrases that compare the subject to something else, suggesting superiority, inferiority, or equality.
Adverbs and Intensifiers Pattern: Spot adverbs or intensifiers that amplify the sentiment of the associated adjective or verb. They often provide a clue about the strength of the sentiment.
]

Question: What is the sentiment of the following sentence?
Sentence: "more than another " best man " clone by weaving a theme throughout this funny film"
Answer:
Step 1: Selected Patterns: Overall Context Pattern and Descriptive Phrases Pattern (Phrases: "weaving a theme" and "funny film").
Step 2: Apply selected patterns to identify the sentiment of the sentence independently.
Step 2.1: Based on Overall Context pattern: The entirety of the sentence portrays a film that is superior and offers added value, emphasizing its distinctiveness and humorous quality. The overall sentiment is positive.
Step 2.2: Based on Descriptive Phrases pattern: The descriptive phrases highlight a positive sentiment in the film being more than just a clone and having a funny theme. The sentiment is positive.
Step 3: Re-evaluate sentiment:
sentiment: positive

Question: What is the sentiment of the following sentence?
Sentence: "contains no wit , only labored gags"
Answer:
Step 1: Selected Patterns: Overall Context Pattern and Negation Handling Pattern (Phrase: "contains no").
Step 2: Apply selected patterns to identify the sentiment of the sentence independently.
Step 2.1: Based on Overall Context pattern: The entire sentence conveys a lack of genuine humor and wit, and instead presents the humor as contrived or forced. The overall sentiment is negative.
Step 2.2: Based on Negation Handling Pattern: The negation "contains no" highlights a lack of wit. It is further emphasized by "labored gags", suggesting forced or contrived humor. Thus, the sentiment is negative regarding the quality or genuineness of the humor.
Step 3: Re-evaluate sentiment:
sentiment: negative
\end{lstlisting}

\subsection{The Demonstration for SNLI Dataset}
\begin{lstlisting}
Please solve the natural language inference task. Specifically, given a premise and a hypothesis, determine the validity of the hypothesis based on the premise:
Yes: The hypothesis is logically derived or directly follows from the premise.
No: The premise provides evidence that refutes the hypothesis.
It is impossible to tell: The premise does not provide sufficient information to confirm or refute the hypothesis.

You will select the most appropriate pattern in the pattern list below to classify the natural language inference task. For each sentence, you will use the selected patterns to identify the relationship between the premise and the hypothesis.


Pattern list:
[
Explicit Evidence Pattern: When the hypothesis directly restates or paraphrases information present in the premise, i.e., premise provides direct evidence that supports the hypothesis, the answer is "yes".
Explicit Contradiction Pattern: The hypothesis contains information that directly negates or opposes a clear statement in the premise. If this condition is met, the answer is "no".
Confident Neutral Pattern: If it is very certain that the hypothesis neither contradicts nor supports the premise in any evident or implicit manner, and the relationship between them is clearly independent, the answer is "it is not possible to tell".
Implicit Contradiction or Neutral Pattern: In this case, no direct contradiction is found. If the hypothesis, when extended logically, negates or conflicts with any part of the premise, even if not directly. If it does, it leans towards contradiction ('no). If no such implicit contradiction is found and the relationship between hypothesis and premise is remains ambiguous, it could be neutral (it is not possible to tell).
Implicit Evidence or Neutral Pattern:  In cases where no direct evidence in the premise supports the hypothesis, the following steps should be applied: Check each element of the hypothesis against the premise. If each element of the hypothesis, when drawing from the premise using world knowledge or logical reasoning, one can infer or reasonably support the entire hypothesis, it leans towards implicit entailment ('yes'). If any part of the hypothesis lacks inferable evidence from the premise or if the connection between the entire hypothesis and premise remains ambiguous, it leans towards neutral ('it is not possible to tell').
Implicit Contradiction or Neutral Pattern: In cases where no direct evidence in the premise negates the hypothesis, the following steps should be applied: Check each element of the hypothesis against the premise. If any element of the hypothesis, when juxtaposed with the premise and utilizing world knowledge or logical reasoning, can subtly negate or contradict any part of the premise, it leans towards implicit contradiction ('no'). If each element of the hypothesis does not provide such subtle contradictions or if the connection between the entire hypothesis and premise remains ambiguous, it leans towards neutral ('it is not possible to tell').
]

Premise: "Children smiling and waving at camera."
Based on this premise, can we conclude the hypothesis "There are children present." is true?
OPTIONS:
- yes
- no
- it is not possible to tell
Answer:
Step 1: Select the most appropriate pattern: Explicit Evidence Pattern
Step 2: Using the Explicit Evidence Pattern, the premise directly states that there are children who are smiling and waving at the camera. This information supports the hypothesis, which claims that there are children present. So, this heuristic suggest a "yes" answer.
Therefore, the answer is: yes

Premise: "A couple play in the tide with their young son."
Based on this premise,can we conclude the hypothesis "The family is sitting down for dinner." is true?
OPTIONS:
- yes
- no
- it is not possible to tell
Answer:
Step 1: Select the most appropriate pattern: Implicit Contradiction or Neutral Pattern
Step 2: Using Implicit Contradiction or Neutral Pattern, although there is no contradiction in the premise describes a scene where a couple and their son are playing in the tide, which suggests they are outdoors and likely at a beach or coastal setting. The hypothesis, on the other hand, presents a scenario where the family is sitting down for dinner, typically an indoor activity. These two activities - playing in the tide and sitting down for dinner are mutually exclusive in the context of a single time frame. Because some element of the hypothesis is identified implicitly contradicts the premise. This heuristic leans towards the answer "no".
Therefore, the answer is: no

Premise: "A young family enjoys feeling ocean waves lap at their feet."
Based on this premise, can we conclude the hypothesis "A young man and woman take their child to the beach for the first time." is true?
OPTIONS:
- yes
- no
- it is not possible to tell
Answer:
Step 1: Select the most appropriate pattern: Implicit Evidence or Neutral Pattern
Step 2: Using Implicit Evidence or Neutral Pattern, "a young family" typically implies the presence of a younger couple and potentially their child, so this part can be inferred. However, the statement about it being the child's "first time" at the beach cannot be inferred or supported from the premise. Because not all elements of the hypothesis can be inferred or supported from the premise, this heuristic leans towards "it is not possible to tell".
Therefore, the answer is: it is not possible to tell.
\end{lstlisting}
\subsection{The Demonstration for RAMS Dataset}
\begin{lstlisting}
Your task is Event Argument Extraction. In this task, you will be provided with a document that describes an event and the goal is to extract the event arguments that correspond to each argument role associated with the event. The terminologies for this task is as follows:
Event trigger: the main word that most clearly expresses an event occurrence, typically a verb or a noun. The trigger word is located between special tokens "<t>" and "<\t>" in the document, and only the event argument explicitly linked to the trigger word should be considered.
Event argument: an entity mention, temporal expression or value that serves as a participant or attribute with a specific role in an event. Event arguments should be quoted exactly as the it appears in the given document.
Argument role: the relationship between an argument to the event in which it participates.
Heuristics: serving as guiding rules for extracting event arguments.

Specifically, you will use the heuristic provided in the heuristic list to guide identify event arguments, and re-evaluate the identified argument candidates to get the final answer.
heuristic list:
[
Semantic Heuristic: [giver] is the person, group, or organization in the document that gives the grant or gift.
Syntactic Heuristic: The [giver] may be recognized by analyzing sentence structure, often appearing before prepositional phrases starting with 'to' that introduce the recipient (e.g., "X gives Y to Z", X is the 'giver').
Dependency Parsing Heuristic: In parsing the sentence structure, the [giver] is often connected through a dependency relation (e.g., 'nsubj' for nominal subject) to the main verb representing the giving action.
]


Example task:
Question: Extract the event arguments of giver, beneficiary, and recipient in the "transaction.transaction.giftgrantprovideaid" event in the provided document, with the trigger word being "granted", highlighted between "<t>" and "</t>". When pinpointing each event argument, it's crucial to quote the entity exactly as it appears in the text. If an event argument is not explicitly mentioned or cannot be directly associated with the event indicated by the trigger word, please respond with "not specified".
Document: a news document
trigger sentence: "The access to the research center in the city was <t>granted</t> by the administrator. The man, Ripley Johnson, earned it."

Answer:
Elaborate the meaning of event type and its argument roles:
"transaction.transaction.giftgrantprovideaid": The event involves a transfer of money or resources in the form of a gift, grant, or provision of aid, signaled by the action of granting.
[giver]: the giver is the person, group, or organization that provides or grants money, resources, or access in the event.
[beneficiary]: the beneficiary is the party who ultimately benefits from the transaction.
[recipient]: the recipient is the entity that receives the money, resources, or access granted in the event.

Recognizing [giver] in the given document:
Step 1: Select one or two heuristics in the heuristic list that are most suitable to identify the [giver] in the given document: Semantic Heuristic and Syntactic Heuristic.
Step 2: Apply selected heuristics to identify [giver] independently.
Step 2.1: Identify the [giver] based on Semantic Heuristic: "[giver] is the person, group, or organization that gives the grant or gift in the document". Applying this heuristic to the document, the entity that gives access of the research center is "administrator".
Step 2.2: Identify the [giver] based on Syntactic Heuristic: "The [giver] may be recognized by analyzing sentence structure, often appearing before prepositional phrases starting with 'to' that introduce the recipient (e.g., 'X gives Y to Z', X is the 'giver')". Applying this heuristic to the given document, the entity that granted access to the research center is 'research center'.
Step 3 Reevaluate argument candidates: ["administrator", "research center"]
Is argument "administrator" alignment with the argument role [giver]? Yes, because "administrator" is directly responsible for the action of granting, establishing their role as the provider of access in the event.
Is argument "research center" alignment with the argument role [giver]? No, because "research center" is the place that access has been granted to, but it doesn't give access.
[giver]: "administrator"

Recognizing [beneficiary] in the given document:
Step 1: Select one or two heuristics in the heuristic list that are most suitable to identify the [beneficiary] in the given document: Semantic Heuristic.
Step 2: Apply selected heuristics to identify [beneficiary] independently.
Step 2.1 Identify the [beneficiary] based on Semantic Heuristic: "[beneficiary] is the entity that ultimately benefits from the gift or grant". Applying this heuristic to the given document, the entity that ultimately benefits from the grant is "not specified".
Step 3 Reevaluate argument candidate: ["not specified"]
Is argument "not specified" alignment with the argument role [beneficiary]? Yes, because the [beneficiary] is not explicitly mentioned so "not specified" is correct.
[beneficiary]: "not specified"

Recognizing [recipient] in the given document:
Step 1: Select one or two heuristics in the heuristic list that are most suitable to identify the [recipient] in the given document: Semantic Heuristic and Dependency Parsing Heuristic.
Step 2: Apply selected heuristics to identify [recipient] independently.
Step 2.2: Identify the [recipient] based on Semantic Heuristic: "[recipient] is the entity that receives the gift or grant". Applying heuristic f1 to the given document, the entity that receives the gift or grant is "Ripley Johnson".
Step 2.1 Identify the [recipient] based on Dependency Parsing Heuristic: "[recipient] is often highlighted in the sentence through a dependency relation that denotes the receiver of the action, such as 'dobj' (direct object) for direct transactions linked to the main verb of the event". Applying this heuristic to the given document, the entity connected to the verb 'granted' through a dobj relation is "Ripley Johnson".
Step 3 Reevaluate argument candidate: ["Ripley Johnson"]
Is argument "Ripley Johnson" alignment with the argument role [recipient]? Yes, because phrase "earned it" implies that "Ripley Johnson" was the intended recipient of the access, aligning with the role of [recipient] in the context of the event.
[recipient]: "Ripley Johnson"


Target task:
\end{lstlisting}

\subsection{The demonstration for DocEE Dataset}

\begin{lstlisting}
Objective: Your task is Event Argument Extraction. In this task, you will be provided with a document that describes an event and the goal is to extract the event arguments that correspond to each argument role associated with the event. The terminologies for this task is as follows:
Key Terminologies:
Event argument: an entity mention, temporal expression or value that serves as a participant or attribute with a specific role in an event. Event arguments should be quoted exactly as the it appears in the given document.
Argument role: the relationship between an argument to the event in which it participates.
Heuristics: serving as guiding principles or strategies to aid the extraction of event arguments, tailored to specific argument roles.

Specifically, you will adapt a set of given heuristics for identifying the argument role of 'giver' to other target argument roles, and then use these adapted heuristics to guide the extraction of target event arguments. Finally, re-evaluate the identified argument candidates to confirm if they are correct event arguments or not.
Heuristic list:
[
Semantic Heuristic: [giver] is the person, group, or organization in the document that gives the grant or gift.
Syntactic Heuristic: The [giver] may be recognized by analyzing sentence structure, often appearing before prepositional phrases starting with 'to' that introduce the recipient (e.g., "X gives Y to Z", X is the 'giver').
Dependency Parsing Heuristic: In parsing the sentence structure, the [giver] is often connected through a dependency relation (e.g., 'nsubj' for nominal subject) to the main verb representing the giving action.
]
When extracting event arguments from the given document, you should also follow the Argument extraction principles below.
Argument Format Principle: articles and prepositions are not included in the identified event argument. For example, the answer should be "damaged car" rather than "damaged car belonging to the victim" or "the damaged car".
Argument Number Principle: In general, each event argument only has a singular answer in the document. However, if and only if, you are highly confident that the mentions associated with an argument role are distinctively different, you may extract no more than three answers for that argument role.


Example sample:
Question: Extract the event arguments of 'Date', 'Casualties and Losses', 'Magnitude', 'Number of Destroyed Building' in the 'Earthquakes' event in the provided news document. When pinpointing each event argument, it's crucial to quote the entity exactly as it appears in the text. Note that if an event argument is not explicitly mentioned or cannot be directly associated with its argument role in question, please respond with "not specified".
Document: a news, the content is omitted here
Answer:
Elaborate the meaning of event type and its argument roles:
'Earthquakes': The event involves the shaking of the surface of the Earth resulting from a sudden release of energy in the Earth's lithosphere.
[Date]: the time when the earthquake occurred.
[Casualties and Losses]: the number of people killed or injured, and the amount of economic losses caused by the earthquake.
[Magnitude]: the measure of the size or intensity of the earthquake.
[Number of Destroyed Building]: the number of buildings or structures that were damaged or destroyed due to the earthquake.

Recognizing [Date] in the given document:
Step 1 Select a heuristic in the heuristic list that is most suitable to identify the [Date] in the given document: Semantic Heuristic.
Step 2 Identify the argument based on Semantic Heuristic: [Date] is the time when the earthquake occurred. Applying this heuristic to the document, the time when the earthquake occurred is "not specified".
Step 3: reevaluate_argument_candidates:
Is argument "not specified" alignment with the argument role [Date]? Yes, because [Date] is not explicitly mentioned in the document, so "not specified" is correct.
[Date]: "not specified"

Recognizing [Casualties and Losses] in the given document:
Step 1 Select a heuristic in the heuristic list that is most suitable to identify the [Casualties and Losses] in the given document: Semantic Heuristic.
Step 2 Identify the argument based on Semantic Heuristic: [Casualties and Losses] is the number of people killed or injured, and the amount of economic losses caused by the earthquake. Applying this heuristic to the document, the [Casualties and Losses] is "claimed 142 deaths" and "800 houses were damaged".
Is argument "claimed 142 deaths" alignment with the argument role [Casualties and Losses]? Yes, because "claimed 142 deaths" provides details about the number of people killed due to the earthquake, aligning with the argument role [Casualties and Losses].
Is argument "800 houses were damaged" alignment with the argument role [Casualties and Losses]? Yes, because it describes the extent of economic losses caused by the earthquake, aligning with the role of [Casualties and Losses].
[Casualties and Losses]: "claimed 142 deaths", "800 houses were damaged"

Recognizing [Magnitude] in the given document:
Step 1 Select a heuristic in the heuristic list that is most suitable to identify the [Magnitude] in the given document: Semantic Heuristic.
Step 2 Identify the argument based on Semantic Heuristic: [Magnitude] is the measure of the size or intensity of the earthquake. Applying this heuristic to the given document, the magnitude of the earthquake is "6.6".
Step 3: reevaluate_argument_candidates:
Is argument "6.6" alignment with the argument role [Magnitude]? Yes, because "The magnitude 6.6 shock" indicate the size of the earthquake.
[Magnitude]: "6.6"

Recognizing [Number of Destroyed Building] in the given document:
Step 1 Select a heuristic in the heuristic list that is most suitable to identify the [Number of Destroyed Building] in the given document: Semantic Heuristic.
Step 2 Identify the argument based on Semantic Heuristic: [Number of Destroyed Building] is the number of buildings or structures that were damaged or destroyed due to the earthquake. Applying this heuristic to the document, the number of buildings or structures that were destroyed due to the earthquake is "more than 80% of the houses".
Step 3: reevaluate_argument_candidates:
Is argument "more than 80% of the houses" alignment with the argument role [Number of Destroyed Building]? No, because the argument describes the percentage of destroyed buildings not the specific number of buildings destroyed.
[Number of Destroyed Building]: "not specified"


Target sample:

\end{lstlisting}

\section{An example of LLM output}
\label{llm_output}
We show an example of the LLM's output of the HD-LoA prompting on the RAMS dataset below:

\begin{lstlisting}
Elaborate the meaning of event type and its argument roles:
"life.die.na": The event involves the death of one or more individuals, signaled by the action of massacre.
[victim]: the victim is the person or group of people who are killed in the event.
[place]: the place is the location where the event occurs.

Recognizing [victim] in the given document:
Step 1: Select one or two heuristics in the heuristic list that are most suitable to identify the [victim] in the given document: Semantic Heuristic.
Step 2: Apply selected heuristics to identify [victim] independently.
Step 2.1: Identify the [victim] based on Semantic Heuristic: "[victim] is the person or group of people who are killed in the event". Applying this heuristic to the document, the victims of the massacre are "280 victims , including women , children and old people".
Step 3 Reevaluate argument candidates: ["280 victims , including women , children and old people"]
Is argument "280 victims , including women , children and old people" alignment with the argument role [victim]? Yes, because "280 victims , including women , children and old people" are the ones who were killed in the massacre, establishing their role as the victims in the event.
[victim]: "280 victims , including women , children and old people"

Recognizing [place] in the given document:
Step 1: Select one or two heuristics in the heuristic list that are most suitable to identify the [place] in the given document: Semantic Heuristic.
Step 2: Apply selected heuristics to identify [place] independently.
Step 2.1: Identify the [place] based on Semantic Heuristic: "[place] is the location where the event occurs". Applying this heuristic to the document, the place where the massacre occurred is "Syria".
Step 3 Reevaluate argument candidates: ["Syria"]
Is argument "Syria" alignment with the argument role [place]? Yes, because "Syria" is the location where the massacre took place, aligning with the role of [place] in the context of the event.
[place]: "Syria"
\end{lstlisting}
\end{document}